\documentclass{article}

\usepackage{microtype}
\usepackage{graphicx}
\usepackage{subcaption}
\usepackage{booktabs} % for professional tables

\usepackage{tabularx}
\usepackage{threeparttable}
\usepackage{multirow}
\usepackage{makecell}
\usepackage[table]{xcolor} % 必须包含 table 选项

\usepackage{enumitem}

\usepackage{hyperref}

\usepackage{algpseudocode}
\usepackage{algorithm}

\usepackage[preprint]{icml2026}

\usepackage{amsmath}
\usepackage{amssymb}
\usepackage{mathtools}
\usepackage{amsthm}
\usepackage{bm}

\usepackage[capitalize,noabbrev]{cleveref}

\theoremstyle{plain}

\theoremstyle{definition}

\theoremstyle{remark}

\usepackage{epigraph}
\usepackage{marvosym}

\usepackage{graphicx}

\usepackage[textsize=tiny]{todonotes}

\usepackage[most]{tcolorbox}
\definecolor{labblue}{HTML}{00AEEF}      % Logo的主色（明亮的蓝色）
\definecolor{lablightblue}{HTML}{F0F7FF} % 极浅的蓝色背景
\definecolor{labdarkblue}{HTML}{0071BC}  % 深蓝色（用于标题或图标）

\newtcolorbox{iclrabstractbox}{
  enhanced,
  colback=lablightblue!0,
  center,
  colframe=labdarkblue!40,
  boxrule=1pt,
  arc=6pt,
  left=0pt,right=0pt,top=8pt,bottom=0pt,
  width=\linewidth,   
}

\icmltitlerunning{SpiralFormer: Looped Transformers Can Learn Hierarchical Dependencies via Multi-Resolution Recursion}

\begin{document}

% \twocolumn[
  % \icmltitle{Submission and Formatting Instructions for \\
  %   International Conference on Machine Learning (ICML 2026)}
\twocolumn[
\icmltitle{SpiralFormer: Looped Transformers Can Learn Hierarchical Dependencies\\
via Multi-Resolution Recursion}

% symbols
\icmlsetsymbol{equal}{\dag}
\icmlsetsymbol{a}{\(\alpha\)}
\icmlsetsymbol{z}{\(\zeta\)}
\icmlsetsymbol{s}{\(\S\)}
\icmlsetsymbol{c}{\Letter} % corresponding author mark

\begin{icmlauthorlist}
\vspace{-12pt}
\large\bfseries
  \icmlauthor{{Chengting Yu}}{equal,z,a}
  \icmlauthor{Xiaobo Shu}{equal,a}
  \icmlauthor{Yadao Wang}{a}
  \icmlauthor{Yizhen Zhang}{a}
  \icmlauthor{Haoyi Wu}{s,a}
  \icmlauthor{You Wu}{s,a}
  \icmlauthor{Rujiao Long}{a}
  \icmlauthor{Ziheng Chen}{a}
  \icmlauthor{Yuchi Xu}{a}
  \icmlauthor{Wenbo Su}{c,a}
  \icmlauthor{Bo Zheng}{a}
\end{icmlauthorlist}

% 机构：手动紧跟在作者行下方显示（按你截图那种）
{\center
% \vspace{2pt}
% \small
% \large
\normalsize\sffamily
\textsuperscript{\(\alpha\)} {Alibaba Group} \quad
\textsuperscript{\(\zeta\)} {Zhejiang University} \quad
\textsuperscript{\(\S\)} {ShanghaiTech University}
\par
}

{
\center
\vspace{-1pt}
\small
\textsuperscript{\dag} Equal contribution \quad 
\textsuperscript{\Letter} Corresponding author
\par
}

% 仍然要定义这些（否则 \printAffiliationsAndNotice 会报错）
% \icmlaffiliation{a}{Alibaba Group}
% \icmlaffiliation{z}{Zhejiang University}
% \icmlaffiliation{s}{ShanghaiTech University}

% \icmlcorrespondingauthor{Bo Zheng}{ }

% 必须调用一次，避免模板警告；这里尽量让脚注里不重复打印机构
% \printAffiliationsAndNotice{%
%   \textsuperscript{\dag}Equal contribution%
% }
% \printAffiliationsAndNotice{\textsuperscript{\dag} Equal contribution \quad \textsuperscript{\Letter} Corresponding author} 

\vskip 0.3in



% \printAffiliationsAndNotice{\textsuperscript{\dag} Equal contribution \quad \textsuperscript{\Letter} Corresponding author}  % no special notice (required even if empty)

\vspace{-15pt}
\begin{iclrabstractbox}
\begin{abstract}
\vspace{14pt}
Recursive (looped) Transformers decouple computational depth from parameter depth by repeatedly applying shared layers, providing an explicit architectural primitive for iterative refinement and latent reasoning.
However, early looped Transformers often underperform non-recursive baselines of equal compute.
While recent literature has introduced more effective recursion mechanisms to mitigate this gap, existing architectures still operate at a fixed, full-token resolution, neglecting the potential efficiency of computing over compressed latent representations.
In this paper, we propose \textbf{SpiralFormer}, a looped Transformer that executes recurrence under a \textbf{multi-resolution recursion} schedule. 
We provide probing evidence that multi-resolution recursion enables the model to learn hierarchical dependencies by inducing iteration-wise functional specialization across different scales. 
Empirically, SpiralFormer achieves better parameter and compute efficiency than both looped and non-looped baselines across model scales from 160M to 1.4B, establishing sequence resolution as a potential axis for scaling recursive architectures.
\end{abstract}

% \vspace{14pt}
% \textbf{Abstract.} Recursive (looped) Transformers decouple computational depth from parameter depth by repeatedly applying shared layers, providing an explicit architectural primitive for iterative refinement and latent reasoning.
% However, early looped Transformers often underperform non-recursive baselines of equal compute.
% While recent literature has introduced more effective recursion mechanisms to mitigate this gap, existing architectures still operate at a fixed, full-token resolution, neglecting the potential efficiency of computing over compressed latent representations.
% In this paper, we propose \textbf{SpiralFormer}, a looped Transformer that executes recurrence under a \textbf{multi-resolution recursion} schedule. 
% We provide probing evidence that multi-resolution recursion enables the model to learn hierarchical dependencies by inducing iteration-wise functional specialization across different scales. 
% Empirically, SpiralFormer achieves better parameter and compute efficiency than both looped and non-looped baselines across model scales from 160M to 1.4B, establishing sequence resolution as a potential axis for scaling recursive architectures.
% \vspace{14pt}
\end{iclrabstractbox}
\vspace{5pt}
]

\section{Introduction}
Scaling model parameters and data has been the dominant driver behind the rapid progress of large language models (LLMs) \citep{brown2020language, chowdhery2023palm}. Yet this axis faces growing headwinds: high-quality text is finite, training and deployment costs are substantial, and memory/communication overheads scale unfavorably with model size \citep{villalobos2022will, patterson2021carbon}. These constraints have motivated efficient architectures that improve \emph{capability per parameter} by 
allocating more computation within 
% increasing compute per parameter under
a fixed parameter budget.

A promising alternative is \textbf{recursive (looped) Transformers}, which decouple computational depth from parameter depth by repeatedly applying a shared set of layers \citep{dehghani2018universal, geiping2025scaling,zhu2025scaling}. Beyond parameter efficiency, looping provides an explicit substrate for \emph{iterative refinement}: the model can update its hidden representation multiple times before emitting the next token, resembling ``thinking time'' in latent space \citep{geiping2025scaling,zeng2025pretraining,zeng2025ponderlm}. This perspective aligns looped Transformers with recent progress on \emph{latent reasoning}, where models perform additional computation in continuous space (e.g., latent thoughts) instead of externalizing every intermediate step as explicit text \citep{hao2024training,zhu2025survey,li2025implicit}.

However, despite this conceptual appeal, early looped Transformers often underperform equally expensive non-recursive baselines. A growing literature mitigates these issues by enriching the recursion mechanism \citep{yu2025mesh, bae2025mixture, wu2025parallel, heo2025ringformer}. While effective, these approaches keep every iteration at the \emph{full-length} sequence, operating at a single, fixed resolution and leaving the resolution axis of recursion largely unexplored
% —despite mounting evidence 
suggesting
from latent reasoning that many reasoning steps can be carried out on compressed representations.

% In contrast, latent reasoning work \citep{hao2024training,cheng2024compressedchainthoughtefficient} hints at a complementary viewpoint: intermediate computation need not operate at full token granularity. Many methods can be interpreted as introducing a limited number of \emph{high-capacity latent slots} that compress multi-token information and support richer computation per step, improving token efficiency \cite{li2025implicit,chen2025reasoning,shi2025swireasoning}. This suggests a missing architectural axis for recursive Transformers: if early iterations are primarily global and low-frequency while later iterations refine local details, then forcing every loop to re-run full-resolution attention may be both inefficient and structurally misaligned with hierarchical reasoning.

Complementarily, latent reasoning work \citep{hao2024training,cheng2024compressedchainthoughtefficient} suggests that intermediate computation need not operate at full token granularity.
Many approaches can be viewed as introducing a limited number of \emph{high-capacity latent slots} (e.g., chunk-level summaries) that compress multi-token information and support richer computation per step, improving token efficiency \citep{li2025implicit,chen2025reasoning,shi2025swireasoning}. This points to a missing architectural axis for recursive Transformers: if loop iterations may involve early global processing followed by later local refinement, then re-running full-resolution attention at every loop may be computationally wasteful and misaligned with a hierarchical dependency structure.

Inspired by the view that latent reasoning benefits from operating over a small set of high-capacity latent slots, we ask whether the same compression form can be made an explicit architectural primitive \emph{within} recursion. We propose \textbf{SpiralFormer}, a looped Transformer that executes a single shared core across a \textbf{multi-resolution recursion} schedule. 
Each iteration constructs a compressed latent sequence by downsampling current hidden states into chunk-level slots, applies the shared Transformer core to capture interactions at that specific resolution, and then causally upscales the result back to token resolution as an update. 
% By varying resolutions across iterations, SpiralFormer induces iteration-wise functional specialization without introducing per-iteration core parameters. 
Empirically, we show that multi-resolution recursion enables looped Transformers to learn hierarchical dependencies. 
Using attention-based probes, we observe coherent shifts in attention statistics across loop iterations and head-level specialization under changes in resolution. 
We also show that SpiralFormer achieves \textbf{superior parameter and compute efficiency} over both looped and non-looped baselines. Our main contributions are:
\begin{itemize}[leftmargin=*, topsep=0pt, itemsep=0pt, parsep=4pt]
% [leftmargin=1.0em,nosep,topsep=0em]
\item We introduce \textbf{multi-resolution recursion}, a novel architectural primitive for looped Transformers that operationalizes multi-token latent compression directly within the recursive process.
\item We provide probing evidence that multi-resolution recursion enables looped Transformers to learn \textbf{hierarchical, scale-dependent dependencies} by exhibiting iteration-wise functional specialization across different resolutions.
\item We show that SpiralFormer achieves superior \textbf{parameter and compute efficiency}, consistently outperforming looped and non-looped baselines across model scales (160M--1.4B) with fewer compute or parameters.
\end{itemize}

% \newpage
\section{Method}
\label{sec:method}

\begin{figure*}[t]
  \centering
  \vspace{-5pt}
  \includegraphics[width=\textwidth, page=1]{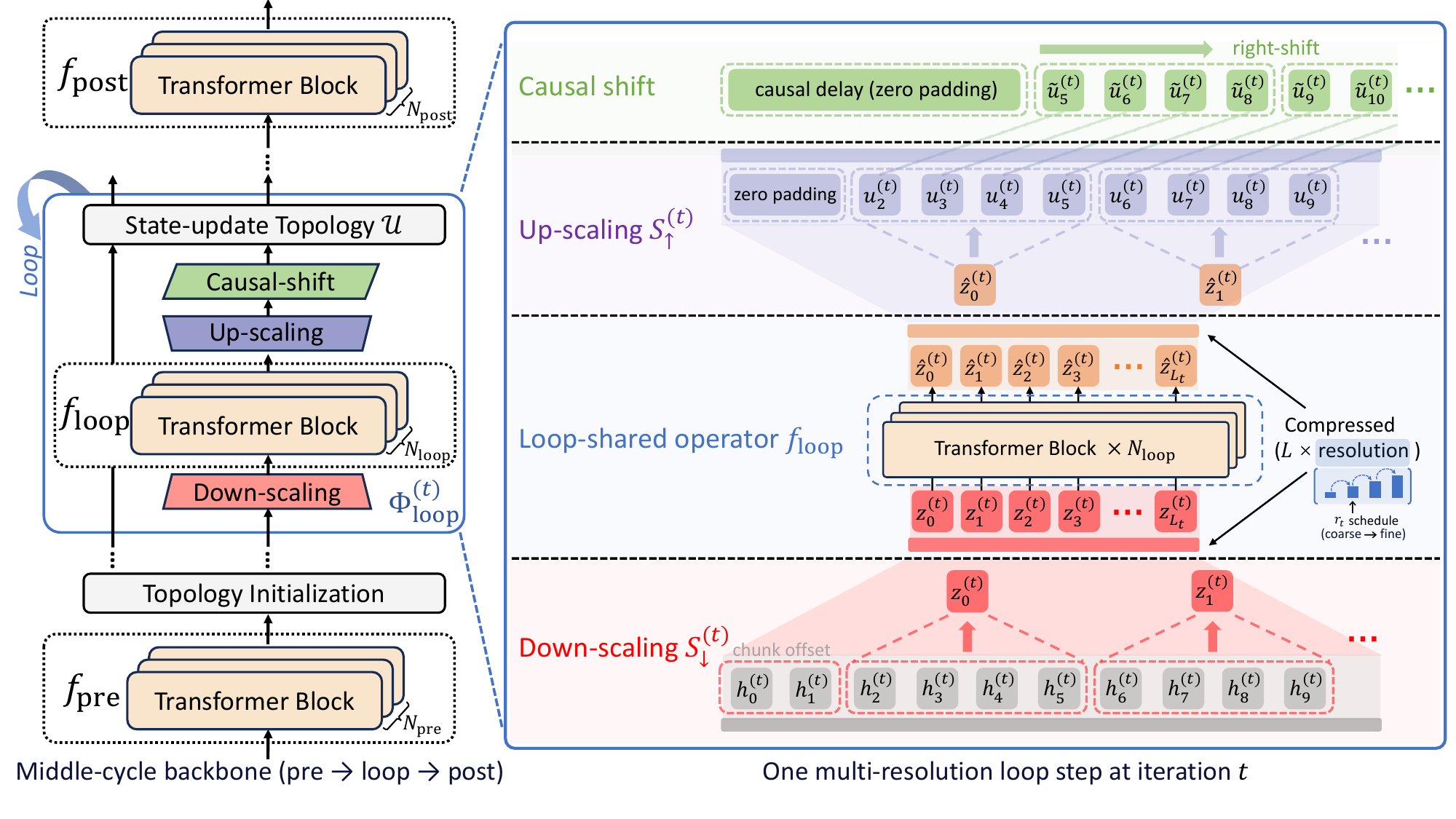}
  \vspace{-25pt}
  \caption{\textbf{SpiralFormer overview.}
\textbf{Left:} we adopt a Middle-cycle backbone (pre→loop→post) where a shared Transformer core is iterated $T$ times (looped recursion) and combined with the running state via a topology update $\mathcal{U}$.
\textbf{Right:} one multi-resolution recursion step at iteration $t$: token-level states $\bm{h}^{(t)}\in\mathbb{R}^{L\times d}$ are causally downsampled into chunk-level latents $\bm{z}^{(t)}\in\mathbb{R}^{L_t\times d}$ (chunk size $g_t = 1/r_t$ with offset $\omega_t$), processed by the shared core to obtain $\widehat{\bm{z}}^{(t)}$, then upsampled back to token-level updates $\bm{u}^{(t)}$.
A causal right-shift by $s_t$ produces $\widetilde{\bm{u}}^{(t)}$, ensuring strict autoregressive causality under compression before updating the loop state.}

\label{fig:spiralformer}
\vspace{-10pt}
\end{figure*}

\subsection{Preliminaries: Recursive (Looped) Transformers}
\label{sec:method-prelim}

% A standard decoder-only Transformer consists of a stack of $L$ distinct layers, where the computational cost and parameter count are coupled. A Recursive (or Looped) Transformer decouples these two by repeatedly applying a shared \textbf{core block} of layers, denoted as $f_{\text{core}}$, for a total of $T$ iterations. This allows for deep computational graphs with a fraction of the parameters.

% A standard decoder-only Transformer consists of a stack of $L$ distinct layers, where the
% computational depth and parameter count are tightly coupled. Recursive (looped) Transformers decouple these two axes by repeatedly applying a shared block for multiple iterations, enabling deeper computation graphs with substantially fewer parameters.

Recursive (looped) Transformers repeatedly apply a loop-shared block for multiple iterations, enabling deeper computation depth with fewer parameters.

\paragraph{Architectural backbone (Middle-cycle).}
We adopt the widely used Middle-cycle architecture
\citep{geiping2025scaling,bae2025mixture,yu2025mesh}.
It consists of (i) a pre-loop block $f_{\mathrm{pre}}$ with $N_{\mathrm{pre}}$ Transformer layers, (ii) a core block
$f_{loop}$ with $N_{\mathrm{loop}}$ layers shared across loop iterations, and (iii) a post-loop block 
$f_{\mathrm{post}}$ with $N_{\mathrm{post}}$ layers.

Let $\bm{x}\in\mathbb{R}^{L\times d}$ denote token embeddings. The model computes
\begin{align}
\bm{h}^{(0)} &= f_{\mathrm{pre}}(\bm{x}), \label{eq:backbone-pre}\\
\bm{h}^{(t+1)} &= \Phi^{(t)}_{\mathrm{loop}}(\bm{h}^{(t)}), \qquad t=0,\dots,T-1, \label{eq:backbone-loop}\\
\bm{h}^{\mathrm{out}} &= f_{\mathrm{post}}(\bm{h}^{(T)}). \label{eq:backbone-post}
\end{align}
% where we use $\Phi^{(t)}_{loop}$ to denote the \emph{entire} loop-step transition that maps $\bm{h}^{(t)}$ to $\bm{h}^{(t+1)}$ (including both the step computation and the topology-dependent state fusion).
where we use $\Phi^{(t)}_{\mathrm{loop}}$ to denote the \emph{entire} loop-step transition that maps $\bm{h}^{(t)}$ to $\bm{h}^{(t+1)}$, combining the computation driven by the shared block $f_{\mathrm{loop}}$ with a state-update topology (defined below).

\paragraph{Topological variants.}
Recursive architectures can differ in how each iteration fuses newly computed
information into the running state. We abstract this by a topology update operator $\mathcal{U}$
that takes (i) the current loop state $\bm{h}^{(t)}$, (ii) a step-produced token-level update
% $\widetilde{\bm{u}}^{(t)}=f_{loop}(\bm{h}^{(t)})$
$\widetilde{\bm{u}}^{(t)}$
(defined in \S\ref{sec:method-multires}), and (iii) an optional topology
carrier $\mathcal{H}^{(t)}$ that stores cross-iteration information, as:
\begin{equation}
(\bm{h}^{(t+1)}, \mathcal{H}^{(t+1)})
=
\mathcal{U}\!\left(\widetilde{\bm{u}}^{(t)},\, \bm{h}^{(t)},\, \mathcal{H}^{(t)};\, t\right).
\label{eq:topo-U}
\end{equation}
% Under the notation, $\Phi^{(t)}$ can be viewed as the composition of the per-step computation $f_{core}(\bm{h}^{(t)})$ that produces $\widetilde{\bm{u}}^{(t)}$ and the topology update $\mathcal{U}$. 
In our experiments we instantiate $\mathcal{U}$ as either Anchor or MeSH, defined below.

% \textbf{Anchor topology.}
% Each iteration is attached to a fixed anchor state $\bm{a}$ ($\bm{a}=\bm{h}^{(0)}$):
% \begin{equation}
% \mathcal{U}_{\text{anchor}}(\bm{h}, \bm{u}) \;=\; \bm{u} + \bm{h}^{(0)}.
% \label{eq:anchor-topology}
% \end{equation}
% Unless stated otherwise, we use anchor topology in the main setting.
\textbf{Anchor topology.}
Each iteration is attached to a fixed anchor state $\bm{h}^{(0)} = f_{\mathrm{pre}}(\bm{x})$.
Under the unified interface, the topology carrier is a constant container
$\mathcal{H}^{(t)} \equiv \bm{h}^{(0)}$, and the topology update is
\begin{equation}
\mathcal{U}_{\text{anchor}}\!\left(\widetilde{\bm{u}}^{(t)}, \bm{h}^{(t)}, \mathcal{H}^{(t)};\, t\right)
\;=\;
\left(\widetilde{\bm{u}}^{(t)} + \bm{h}^{(0)},\; \bm{h}^{(0)}\right).
\label{eq:anchor-topology}
\end{equation}

\textbf{MeSH topology.}
MeSH replaces direct state passing with an explicit multi-slot memory buffer and lightweight
step-wise read/write routers \citep{yu2025mesh}. Under MeSH, the topology carrier is
$\mathcal{H}^{(t)}\equiv \bm{M}^{(t)}=\{\bm{m}^{(t)}_b\}_{b=0}^{B-1}$, where each memory slot
$\bm{m}^{(t)}_b\in\mathbb{R}^{L\times d}$ has the same shape as the loop state.
At each iteration $t$, MeSH (i) \emph{writes} the causality-corrected update $\widetilde{\bm{u}}^{(t)}$
into the buffer using a learned write router, and (ii) \emph{reads} from the updated buffer using a
learned read router to synthesize the next loop state $\bm{h}^{(t+1)}$:
\begin{equation}
(\bm{h}^{(t+1)}, \bm{M}^{(t+1)})
=
\mathcal{U}_{\mathrm{mesh}}\!\left(\widetilde{\bm{u}}^{(t)},\, \bm{h}^{(t)},\, \bm{M}^{(t)};\, t\right).
\label{eq:mesh-U-compact}
\end{equation}
We follow \citet{yu2025mesh} and use a transitional write--read initialization to reorganize the pre-loop output into the buffer before entering the main loop. Full update equations are provided in Appendix~\ref{sec:appendix-mesh}.

\subsection{Multi-Resolution Recursion: Executing One Shared Core Across Scales}
\label{sec:method-multires}

Our main idea is to execute different loop iterations at different sequence \emph{resolutions}. Early iterations compute on a short (coarse) sequence to cheaply build global interactions; later iterations progressively increase resolution to refine token-level representations. This produces iteration-wise functional specialization without adding per-iteration core parameters.

\paragraph{Resolution schedule.}
We define a resolution schedule $\{r_t\}_{t=0}^{T-1}$ with $r_t\in(0,1]$, where iteration $t$ uses an effective length
\begin{equation}
% L_t = \lfloor L/g_t\rfloor.
L_t = \lfloor r_t L\rfloor.
\label{eq:Lt}
\end{equation}

\paragraph{One iteration as downscale $\rightarrow$ core $\rightarrow$ upscale.}
At loop iteration $t$, we compute an update by:
\begin{align}
\bm{z}^{(t)} \;&=\; \mathcal{S}^{(t)}_{\downarrow}\!\left(\bm{h}^{(t)}; r_t\right)\in\mathbb{R}^{L_t\times d},
\label{eq:downscale}\\
\widehat{\bm{z}}^{(t)} \;&=\; f_{\mathrm{loop}}\!\left(\bm{z}^{(t)}\right)\in\mathbb{R}^{L_t\times d},
\label{eq:core-on-z}\\
\bm{u}^{(t)} \;&=\; \mathcal{S}^{(t)}_{\uparrow}\!\left(\widehat{\bm{z}}^{(t)}; \bm{h}^{(t)}, r_t\right)\in\mathbb{R}^{L\times d}.
\label{eq:upscale}
\end{align}
We then apply a causality-correcting right-shift operator (defined in \S\ref{sec:method-shift}) to obtain $\widetilde{\bm{u}}^{(t)}$ and update the loop state:
% \begin{equation}
% \bm{h}^{(t+1)} = \mathcal{U}\!\left(\bm{h},\, \widetilde{\bm{u}}^{(t)}\right).
% \label{eq:update}
% \end{equation}
\begin{equation}
(\bm{h}^{(t+1)}, \mathcal{H}^{(t+1)})
=
\mathcal{U}\!\left(\widetilde{\bm{u}}^{(t)},\, \bm{h}^{(t)},\, \mathcal{H}^{(t)};\, t\right).
\label{eq:update}
\end{equation}
% This construction reduces attention cost from $O(TL^2)$ to $O(\sum_tL_t^2)$ as well as mlp cost from $O(TL)$ to $O(\sum_t L_t)$ (plus a small linear overhead for down/up-scaling).

\subsection{Causal Down-scaling and Up-scaling Operators}
\label{sec:operators}

The design of the $\mathcal{S}_{\downarrow}^{(k)}$ and $\mathcal{S}_{\uparrow}^{(k)}$ operators is critical for maintaining the \emph{strict causality} required by autoregressive models: the update written to position $i$ must not depend on tokens $>i$. The main challenge is that chunk-level aggregation summarizes an entire chunk, which includes future tokens relative to earlier positions in the same chunk.
We enforce strict causality by a right-shift that also induces an interpretable overlap structure.

% \subsubsection{Chunking and position maps}
% \label{sec:method-chunking}

% % At iteration $T$, we partition the sequence of length $L$ into non-overlapping chunks of size $g_t = 1/r_t$. For simplicity, we assume $L$ is divisible by $g_k$. 
% For clarity, we focus on a block-wise downsampling scheme where $L_t = L/g_t$ and $g_t\in\mathbb{N}$ is the chunk size at iteration $t$ (assuming divisibility for simplicity).
% We define the \emph{chunk index} and \emph{in-chunk offset} maps:
% \begin{align*}
% &\pi_t(i)=\left\lfloor \frac{i}{g_t}\right\rfloor \in \{0,\dots,L_t-1\},\qquad \\
% &\rho_t(i)= i - g_t\pi_t(i)\in \{0,\dots,g_t-1\}.
% \label{eq:pi-rho}
% \end{align*}
% We also define the chunk position set as
% \begin{equation}
% \mathcal{I}_{t,j}=\{i\in\{0,\dots,L-1\}:\pi_t(i)=j\}.
% \label{eq:chunk-set}
% \end{equation}

\subsubsection{Chunking and position maps}
\label{sec:method-chunking}

At iteration $t$, we downscale the length-$L$ sequence using blockwise chunking with chunk size $g_t = \lfloor \frac{1}{r_t} \rfloor\in\mathbb{N}$.
We introduce an integer \emph{chunk offset} $\omega_t\in\{0,1,\dots,g_t-1\}$, which shifts the chunk boundaries.
We define the (offset) \emph{chunk index map} and \emph{in-chunk map} as
\begin{equation}
\pi_t(i)=\left\lfloor \frac{i + \omega_t}{g_t}\right\rfloor,\qquad
\rho_t(i)= (i + \omega_t) - g_t\pi_t(i),
\label{eq:pi-rho-offset}
\end{equation}
where $i\in\{0,\dots,L-1\}$ and $\rho_t(i)\in\{0,\dots,g_t-1\}$.

\paragraph{Dropping the last incomplete chunk.}
We realize the target resolution $r_t$ using an integer chunk size
$g_t=\lfloor 1/r_t\rfloor$, which yields an effective number of chunks
\begin{equation}
L_t \;=\; \left\lfloor \frac{L}{g_t}\right\rfloor \;\approx\; \lfloor r_t L\rfloor.
\label{eq:Lt-floor}
\end{equation}
The shifted partition may produce an incomplete last chunk; we simply drop it.
Equivalently, we only keep positions
\begin{equation}
\mathcal{V}_t \;=\; \{\, i\in\{0,\dots,L-1\} : \pi_t(i)\le L_t-1 \,\},
\label{eq:valid-positions}
\end{equation}
and define the $j$-th chunk as
\begin{equation}
\mathcal{I}_{t,j}=\{i\in\mathcal{V}_t:\pi_t(i)=j\},\quad j=0,\dots,L_t-1.
\label{eq:chunk-set}
\end{equation}

\paragraph{Default choice.}
Unless otherwise specified, we use the half-chunk offset $\omega_t=\lfloor g_t/2\rfloor$.
We provide further motivation and analysis in Appendix~\ref{sec:appendix-chunk-offset}.

\begin{algorithm}[t]
\caption{SpiralFormer: Multi-Resolution Recursion}
\label{alg:mrr}
\begin{algorithmic}[1]
\Require token embeddings $\bm{x}\in\mathbb{R}^{L\times d}$; pre-loop $f_{\mathrm{pre}}$; loop-shared core $f_{\mathrm{loop}}$; post-loop $f_{\mathrm{post}}$;
schedule $\{r_t\}_{t=0}^{T-1}$; shift sizes $\{s_t\}_{t=0}^{T-1}$; topology update $\mathcal{U}$
\State $\bm{v} \gets f_{\mathrm{pre}}(\bm{x})$
\State $(\bm{h}^{(0)}, \mathcal{H}^{(0)}) \gets \textsc{InitTopo}(\bm{x}, \bm{v})$
\For{$t=0$ to $T-1$}
    \State $\bm{z}^{(t)} \gets \mathcal{S}^{(t)}_{\downarrow}(\bm{h}^{(t)}; r_t)$ \Comment{Down-scaling}
    \State $\widehat{\bm{z}}^{(t)} \gets f_{\mathrm{loop}}(\bm{z}^{(t)})$ \Comment{Shared operator}
    \State $\bm{u}^{(t)} \gets \mathcal{S}^{(t)}_{\uparrow}(\widehat{\bm{z}}^{(t)}; \bm{h}^{(t)}, r_t)$ \Comment{Up-scaling}
    \State $\widetilde{\bm{u}}^{(t)} \gets \textsc{CausalShift}(\bm{u}^{(t)}, s_t)$ \Comment{Causal shift}
    \State $(\bm{h}^{(t+1)}, \mathcal{H}^{(t+1)}) \gets \mathcal{U}(\widetilde{\bm{u}}^{(t)}, \bm{h}^{(t)}, \mathcal{H}^{(t)}; t)$
\EndFor
\State $\bm{h}^{\mathrm{out}} \gets f_{\mathrm{post}}(\bm{h}^{(T)})$
\State \Return $\bm{h}^{\mathrm{out}}$
\end{algorithmic}
\end{algorithm}

\subsubsection{Down-scaling operators $\mathcal{S}^{(t)}_{\downarrow}$}
\label{sec:method-downscale}

\textbf{Mean pooling.}
A simple choice is mean pooling within each chunk:
\begin{equation}
\bm{z}^{(t)}_j \;=\; \frac{1}{g_t}\sum_{i\in\mathcal{I}_{t,j}} \bm{h}^{(t)}_i.
\label{eq:mean-pooling}
\end{equation}

\textbf{Self-aggregation (default).}
Instead, to create the latent vector $\bm{z}_j^{(t)}$, we apply a learnable weighted combination of states within the chunk. A lightweight, iteration-specific scorer $\mathcal{A}^{(t)}: \mathbb{R}^d \to \mathbb{R}$ (one linear layer in our implementation) computes importance weights $\bm{\alpha}^{(k)}$ via Softmax normalization:
\begin{equation}
    \alpha^{(t)}_{j,i}=\mathrm{Softmax}_{i\in\mathcal{I}_{t,j}}\!\big(\mathcal{A}^{(t)}(\bm{h}^{(t)}_i)\big)
    \label{eq:self-agg}
\end{equation}
The latent vector is then the aggregated sum:
\begin{equation}
    \bm{z}^{(t)}_j=\sum_{i\in\mathcal{I}_{t,j}}\alpha^{(t)}_{j,i}\,\bm{h}^{(t)}_i
\label{eq:downscale-general}
\end{equation}
% We learn a within-chunk mixture:
% \begin{align*}
% &\bm{z}^{(t)}_j=\sum_{i\in\mathcal{I}_{t,j}}\alpha^{(t)}_{j,i}\,\bm{h}^{(t)}_i,
% \label{eq:self-agg}
% \end{align*}
% where $s^{(t)}:\mathbb{R}^d\to\mathbb{R}$ is a lightweight scorer (one linear layer in our implementation).

% \subsubsection{Up-scaling operators $\mathcal{S}^{(t)}_{\uparrow}$}
% \label{sec:method-upscale}

% Given $\widehat{\bm{z}}^{(t)}\in\mathbb{R}^{L_t\times d}$, up-scaling writes updates back to fine positions.

% \textbf{Uniform broadcast.} A simple choice is to broadcast within each chunk:
% \begin{equation}
% \bm{u}^{(t)}_i = \widehat{\bm{z}}^{(t)}_{\pi_t(i)}.
% \label{eq:broadcast}
% \end{equation}

% \textbf{Output-dependent allocation (default).}
% Instead, we apply an output-dependent router $\mathcal{B}^{(t)}: \mathbb{R}^d \to \mathbb{R}^{g_t}$ to predict allocation weights $\bm{\beta}^{(t)}$ (one linear layer with softmax in our implementation):
% \begin{equation}
% \bm{\beta}^{(t)}_j=\mathrm{Softmax}\!\left(\mathcal{B}^{(t)}(\widehat{\bm{z}}^{(t)}_j)\right)\in\mathbb{R}^{g_t}
% \end{equation}
% The update for each position $i \in \mathcal{I}_{t,j}$ is then a scaled version of the latent vector:
% \begin{equation}
%     \bm{u}^{(t)}_i = \beta^{(t)}_{\pi_t(i),\rho_t(i)}\cdot \widehat{\bm{z}}^{(t)}_{\pi_t(i)}
% \end{equation}

\subsubsection{Up-scaling operators $\mathcal{S}^{(t)}_{\uparrow}$}
\label{sec:method-upscale}

Given $\widehat{\bm{z}}^{(t)}\in\mathbb{R}^{L_t\times d}$, up-scaling writes updates back to token-level
positions. We use an allocation vector
$\bm{\beta}^{(t)}_j\in\mathbb{R}^{g_t}$ together with a resolution-dependent gain $\lambda_t$ for each valid position $i\in\mathcal{V}_t$:
\begin{equation}
\bm{u}^{(t)}_i
=
\lambda_t\, \beta^{(t)}_{\pi_t(i),\rho_t(i)}\cdot \widehat{\bm{z}}^{(t)}_{\pi_t(i)}.
\label{eq:upscale-general}
\end{equation}

\textbf{Uniform broadcast.}
A simple choice is to broadcast within each chunk with setting $\beta^{(t)}_{j,\rho}=1/g_t$, yielding
\begin{equation}
\bm{u}^{(t)}_i=\frac{\lambda_t}{g_t}\,\widehat{\bm{z}}^{(t)}_{\pi_t(i)}.
\label{eq:upscale-uniform}
\end{equation}

\textbf{Output-dependent allocation (default).}
Instead, we apply an output-dependent router $\mathcal{B}^{(t)}: \mathbb{R}^d \to \mathbb{R}^{g_t}$ to predict allocation weights $\bm{\beta}^{(t)}$ (one linear layer):
\begin{equation}
\bm{\beta}^{(t)}_j
=
\mathrm{Softmax}\!\left(\mathcal{B}^{(t)}(\widehat{\bm{z}}^{(t)}_j)\right)\in\mathbb{R}^{g_t},
\quad
\sum_{\rho=0}^{g_t-1} \beta^{(t)}_{j,\rho}=1.
\label{eq:upscale-router}
\end{equation}
% The update for each position $i \in \mathcal{I}_{t,j}$ is then a scaled version of the latent vector:
% \begin{equation}
% \bm{u}^{(t)}_i
% =
% \lambda_t\, \beta^{(t)}_{\pi_t(i),\rho_t(i)}\cdot \widehat{\bm{z}}^{(t)}_{\pi_t(i)}.
% \end{equation}

\textbf{Gain scaling.} We set the up-scaling gain as $\lambda_t=\sqrt{g_t}$ to calibrate the token-level update magnitude across different chunk sizes.

% We predict per-chunk allocation weights from $\widehat{\bm{z}}^{(t)}_j$:
% \begin{align*}
% &\boldsymbol{\beta}^{(t)}_j=\mathrm{Softmax}\!\left(W^{(t)}\widehat{\bm{z}}^{(t)}_j\right)\in\mathbb{R}^{g_t},
% \qquad \\
% &\bm{u}^{(t)}_i = \beta^{(t)}_{\pi_t(i),\rho_t(i)}\cdot \widehat{\bm{z}}^{(t)}_{\pi_t(i)}.
% \label{eq:output-dependent-up}
% \end{align*}

% \textbf{Input-dependent allocation.}
% Alternatively, $\boldsymbol{\beta}^{(t)}_j$ can be predicted from fine-grained chunk states $\{\bm{h}^{(t)}_i\}_{i\in\mathcal{I}_{t,j}}$; we evaluate this variant in ablations.

\subsubsection{Right-shift for strict causality and overlap regimes}
\label{sec:method-shift}

A naive use of $\bm{u}^{(t)}$ would violate causality, as the representation $\hat{\bm{z}}_j^{(t)}$ depends on the entire chunk $\pi_t(j)$.
We therefore apply a \textbf{causal right-shift} of size $s_t$ to the update tensor, resulting in the final update $\tilde{\bm{u}}^{(k)}$:
% Even if $f_{\mathrm{core}}$ is causal on the compressed sequence, $\bm{u}^{(t)}$ may still leak future information within a chunk because $\widehat{\bm{z}}^{(t)}_j$ summarizes the entire chunk.
% We therefore apply a right shift by $s_t$ positions:
\begin{equation}
\widetilde{\bm{u}}^{(t)}[i] =
\begin{cases}
0, & i < s_t,\\
\bm{u}^{(t)}[i-s_t], & i\ge s_t,
\end{cases}
\qquad s_t\in\mathbb{N}.
\label{eq:right-shift}
\end{equation}
In our main model we set $s_t=g_t-1$, which yields a \emph{single-token overlap} between the chunk generating the update and the chunk receiving it (see motivation in Appendix~\ref{sec:appendix-proof-prop1}).
We also discuss an alternative no-overlap setting ($s_t \ge g_t$) that can be exploited for inference-time pipelining in Appendix~\ref{sec:appendix-parallel-regime}.

\subsection{Default setting of SpiralFormer}

\textbf{Resolution scheduling (Coarse-to-Fine).}
The resolution schedule \(\{r_t\}_{t=0}^{T-1}\) specifies the effective sequence length
\(L_t = \lfloor r_t L \rfloor\) used at iteration \(t\) (Eq.~\eqref{eq:Lt}), and the
chunk size \(g_t\) in our blockwise implementation (Eq.~\eqref{eq:Lt-floor}). Each loop applies the
same shared core \(f_{\mathrm{loop}}\) to a sequence of length \(L_t\), so changing \(\{r_t\}\)
controls the scale at which the core operates. By default, we adopt a \textbf{\emph{coarse-to-fine}} schedule, implemented as a monotone increasing
sequence
\[
r_t = 2\,r_{t-1},
\]
starting from a small initial resolution (e.g., \(r_0 = 1/8\) or \(1/16\)). This realizes progressive refinement: early iterations operate on short latent
sequences that compress many tokens at low cost, while later iterations
run at higher resolutions and refine token-level details.

\textbf{Causality and chunking.}
For each iteration $t$, we ensure strict autoregressive causality by applying the right-shift
operator (Eq.~\eqref{eq:right-shift}) to the upscaled update. By default, we use
$s_t=g_t-1$, the smallest shift that guarantees causality under within-chunk aggregation and
yields a single-token overlap between the chunk producing the update and the chunk receiving it
(Proposition~1, Appendix~\ref{sec:appendix-proof-prop1}).
Unless otherwise stated, we set the chunk offset to $\omega_t=\lfloor g_t/2\rfloor$
(Eq.~\eqref{eq:pi-rho-offset}), which shifts chunk boundaries and modifies the decoding-time
triggering pattern (Appendix~\ref{sec:appendix-chunk-offset} \& Algorithm~\ref{alg:spiralformer-decode-v3}).
% ; see also Appendix
% Algorithm~\ref{alg:spiralformer-decode-v3}).

% A training-time instantiation of multi-resolution recursion over a full sequence is summarized in Algorithm~\ref{alg:mrr}. A token-wise autoregressive decoding procedure, including chunk-triggered multi-resolution updates and causal right-shift under KV caching, is given in Algorithm~\ref{alg:spiralformer-decode-v3} (Appendix~\ref{sec:appendix-exp}).

% ---------- 修改后的 PREAMBLE (请确保包含以下内容) ----------
\newcommand{\anchormark}{\textsuperscript{*}}
\newcommand{\meshmark}{\textsuperscript{\dag}}
\definecolor{colorbetter}{rgb}{0.88, 1.0, 0.88} % 定义浅绿色
% ------------------------------------------------------
\begin{table*}[t]
\caption{
% Main experimental results comparing SpiralFormer against full-resolution baselines and standard looped Transformers (LoopedFormer) across various model sizes. \anchormark Anchor topology. \meshmark MeSH topology. Highlighted values indicate reduction in parameters or FLOPs compared to the baseline.
Main results on Pythia-suite (160M--1.4B) comparing SpiralFormer with (i) the non-recursive \textsc{Baseline (Pythia)} and (ii) full-resolution \textsc{LoopedFormer}.
We report total / non-embedding parameters, prefill FLOPs for 4096-token, validation perplexity and downstream accuracy (0-shot / 5-shot).
Shaded cells indicate reduced parameters or FLOPs relative to the \textsc{Baseline} at the same scale.
\anchormark denotes Anchor topology and \meshmark denotes MeSH topology.
Best results in each scale are in \textbf{bold}, and second-best are \underline{underlined}.
}
\vspace{-15pt}
\label{tab:main_results}
\vskip 0.15in
\centering
\footnotesize
\setlength{\tabcolsep}{2.5pt}
\setlength{\fboxsep}{1.5pt} % 控制高亮色块的边距，使其紧凑地包裹数字
\begin{sc}
\resizebox{\textwidth}{!}{%
\begin{tabular}{ll rc cccc cc}
% \toprule
% \textbf{Model} & \textbf{Config} &
% \multicolumn{1}{c}{\begin{tabular}{@{}c@{}}\\ \textbf{Params (M)}\\(Total/Non-Emb)\end{tabular}} &
% \multicolumn{1}{c}{\begin{tabular}{@{}c@{}}\\ \textbf{FLOPs (1e12)}\\(4096 Prefill)\end{tabular}} &
% \multicolumn{4}{c}{\textbf{Perplexity ↓}} &
% \multicolumn{2}{c}{\textbf{Task Acc ↑}} \\
% \cmidrule(lr){5-8} \cmidrule(lr){9-10}
% &  & & &
% \multicolumn{1}{c}{Pile} &
% \multicolumn{1}{c}{Wiki} &
% \multicolumn{1}{c}{LD-O} &
% \multicolumn{1}{c}{LD-S} &
% \multicolumn{1}{c}{0-shot} &
% \multicolumn{1}{c}{5-shot} \\
% \midrule
\toprule
% 第一行表头
\multirow{2}{*}{\textbf{Model}} & 
\multirow{2}{*}{\textbf{Config}} & 
\multirow{3}{*}{\makecell[b]{\textbf{Params (M)}\\\scriptsize (Total/Non-Emb)}} & 
\multirow{3}{*}{\makecell[b]{\textbf{FLOPs (1e12)}\\\scriptsize (4096 Prefill)}} & 
\multicolumn{4}{c}{\textbf{Perplexity $\downarrow$}} & 
\multicolumn{2}{c}{\textbf{Task Acc $\uparrow$}} \\
\cmidrule(lr){5-8} \cmidrule(lr){9-10}
% 第二行表头
& & & & Pile & Wiki & LD-O & LD-S & 0-shot & 5-shot \\
\midrule

\multicolumn{10}{l}{\emph{Pythia-160M}} \\
\midrule
Baseline (Pythia) & 12 Layers & 163.5 / 85.1 & 1.65 &
11.31 & 30.32 & \underline{42.86} & 175.62 &
\textbf{39.88} & 40.54 \\
LoopedFormer\anchormark & \texttt{2+4×\{1,1\}+2} & \colorbox{colorbetter}{135.2 / 56.7} & 1.65 &
11.63 & 31.69 & 50.38 & 195.11 &
38.81 & 40.15 \\
LoopedFormer\meshmark & \texttt{2+4×\{1,1\}+2} & \colorbox{colorbetter}{135.2 / 56.7} & 1.65 &
11.37 & 30.43 & 46.60 & 178.77 &
39.41 & 40.60 \\
\textbf{SpiralFormer-B}\meshmark & \texttt{2+4×\{$\tfrac{1}{8}$,$\tfrac{1}{4}$,$\tfrac{1}{2}$,1\}+2} & \colorbox{colorbetter}{135.2 / 56.8} & \colorbox{colorbetter}{1.48} &
\underline{11.29} & \underline{30.27} & 43.27 & \underline{155.78} &
\underline{39.73} & \underline{41.02} \\
\textbf{SpiralFormer-L}\meshmark & \texttt{4+4×\{$\tfrac{1}{16}$,$\tfrac{1}{8}$,$\tfrac{1}{4}$,$\tfrac{1}{2}$\}+4} & 163.6 / 85.1 & \colorbox{colorbetter}{1.49} &
\textbf{10.94} & \textbf{28.85} & \textbf{41.24} & \textbf{147.52} &
39.30 & \textbf{41.37} \\
\midrule

\multicolumn{10}{l}{\emph{Pythia-410M}} \\
\midrule
Baseline (Pythia) & 24 Layers & 407.4 / 302.3 & 4.59 &
9.07 & 21.79 & \underline{19.48} & 65.86 &
43.87 & 45.31 \\
LoopedFormer\anchormark & \texttt{4+8×\{1,1\}+4} & \colorbox{colorbetter}{306.7 / 201.5} & 4.59 &
9.19 & 22.12 & 20.37 & 52.55 &
43.70 & 45.68 \\
LoopedFormer\meshmark & \texttt{4+8×\{1,1\}+4} & \colorbox{colorbetter}{306.7 / 201.6} & 4.59 &
9.09 & 21.84 & 19.63 & \underline{42.51} &
44.12 & 45.56 \\
\textbf{SpiralFormer-B}\anchormark & \texttt{4+8×\{$\tfrac{1}{8}$,$\tfrac{1}{4}$,$\tfrac{1}{2}$,1\}+4} & \colorbox{colorbetter}{306.7 / 201.6} & \colorbox{colorbetter}{4.10} &
9.13 & 22.04 & 21.96 & 47.33 &
43.87 & 46.30 \\
\textbf{SpiralFormer-B}\meshmark & \texttt{4+8×\{$\tfrac{1}{8}$,$\tfrac{1}{4}$,$\tfrac{1}{2}$,1\}+4} & \colorbox{colorbetter}{306.8 / 201.6} & \colorbox{colorbetter}{4.11} &
\underline{9.00} & \underline{21.48} & \textbf{19.11} & \textbf{39.78} &
\underline{44.31} & \underline{46.75} \\
\textbf{SpiralFormer-L}\meshmark & \texttt{8+8×\{$\tfrac{1}{16}$,$\tfrac{1}{8}$,$\tfrac{1}{4}$,$\tfrac{1}{2}$\}+8} & 407.5 / 302.4 & \colorbox{colorbetter}{4.16} &
\textbf{8.73} & \textbf{20.55} & 20.38 & 47.89 &
\textbf{44.97} & \textbf{47.06} \\
\midrule

\multicolumn{10}{l}{\emph{Pythia-1B}} \\
\midrule
Baseline (Pythia) & 16 Layers & 1020.2 / 805.7 & 9.67 &
7.96 & 17.66 & 13.53 & 33.65 &
46.95 & 49.07 \\
LoopedFormer\anchormark & \texttt{3+5×\{1,1\}+3} & \colorbox{colorbetter}{768.4 / 553.9} & 9.67 &
8.10 & 18.15 & 13.32 & 32.34 &
46.73 & 48.83 \\
LoopedFormer\meshmark & \texttt{3+5×\{1,1\}+3} & \colorbox{colorbetter}{768.4 / 554.0} & 9.67 &
{7.90} & {17.54} & {12.19} & {26.71} &
{47.53} & {49.51} \\
\textbf{SpiralFormer-B}\meshmark & \texttt{3+5×\{$\tfrac{1}{8}$,$\tfrac{1}{4}$,$\tfrac{1}{2}$,1\}+3} & \colorbox{colorbetter}{768.6 / 554.1} & \colorbox{colorbetter}{8.95} &
\underline{7.80} & \underline{17.21} & \underline{11.96} & \underline{25.55} &
\underline{48.14} & \underline{50.25} \\
\textbf{SpiralFormer-L}\meshmark & \texttt{5+6×\{$\tfrac{1}{16}$,$\tfrac{1}{8}$,$\tfrac{1}{4}$,$\tfrac{1}{2}$\}+5} & 1020.4 / 805.9 & \colorbox{colorbetter}{8.96} &
\textbf{7.64} & \textbf{16.73} & \textbf{11.94} & \textbf{23.90} &
\textbf{48.97} & \textbf{51.83} \\
\midrule

\multicolumn{10}{l}{\emph{Pythia-1.4B}} \\
\midrule
Baseline (Pythia) & 24 Layers & 1423.0 / 1208.6 & 14.08 &
7.44 & 15.97 & 10.51 & 22.81 &
49.50 & 51.93 \\
Baseline\meshmark & 24 Layers & 1423.1 / 1208.7 & 14.08 &
\underline{7.26} & \underline{15.25} & \underline{9.46} & 16.31 &
50.21 & 53.12 \\
LoopedFormer\anchormark & \texttt{4+8×\{1,1\}+4} & \colorbox{colorbetter}{1020.2 / 805.7} & 14.08 &
7.51 & 16.25 & 10.71 & 19.37 &
49.39 & 51.27 \\
LoopedFormer\meshmark & \texttt{4+8×\{1,1\}+4} & \colorbox{colorbetter}{1020.2 / 805.8} & 14.08 &
7.39 & 15.84 & 9.72 & 19.39 &
50.56 & 52.79 \\
\textbf{SpiralFormer-B}\meshmark & \texttt{4+8×\{$\tfrac{1}{8}$,$\tfrac{1}{4}$,$\tfrac{1}{2}$,1\}+4} & \colorbox{colorbetter}{1020.4 / 805.9} & \colorbox{colorbetter}{12.92} &
7.30 & 15.61 & \textbf{9.06} & \underline{15.30} &
\underline{51.48} & \underline{53.22} \\
\textbf{SpiralFormer-L}\meshmark & \texttt{8+8×\{$\tfrac{1}{16}$,$\tfrac{1}{8}$,$\tfrac{1}{4}$,$\tfrac{1}{2}$\}+8} & 1423.2 / 1208.8 & \colorbox{colorbetter}{13.13} &
\textbf{7.14} & \textbf{15.03} & 9.73 & \textbf{14.42} &
\textbf{51.75} & \textbf{54.37} \\
\bottomrule
\end{tabular}
}
\end{sc}
\vspace{-10pt}
\end{table*}

\section{Experiments}
\label{sec:experiments}

\subsection{Setup}
\label{sec:exp-setup}

We evaluate SpiralFormer by pretraining decoder-only Transformers on the Pythia family (160M--1.4B) \citep{biderman2023pythia}.
% and comparing against both non-recursive and recursive baselines under matched parameter and/or compute budgets. 
All models are trained from scratch on a deduplicated subset of the Pile \citep{gao2020pile} for one epoch (250B tokens) and are evaluated by (i) language modeling perplexity and (ii) few-shot downstream accuracy. Additional implementation details are provided in Appendix~\ref{sec:appendix-exp}.
% (training hyperparameters, evaluation protocol, and compute accounting) are provided in Appendix~\ref{sec:appendix-exp}.

\textbf{Models and configurations.}
We evaluate three model families at each Pythia scale: (i) the \emph{standard} Pythia \textsc{Baseline}; (ii) the \emph{full-resolution recursive} \textsc{LoopedFormer} that matches the unrolled compute depth by looping a shared core; and (iii) \textsc{SpiralFormer}, which replaces each loop step by \emph{multi-resolution recursion}. We denote recursive layer allocations as \texttt{$N_{\mathrm{pre}}$+$N_{\mathrm{loop}}$×\{$r_0,\dots,r_{T-1}$\}+$N_{\mathrm{post}}$}, where \(\{r_t\}\) is the resolution schedule of the \(T\) loop iterations (\S\ref{sec:method-multires}). In particular, \textsc{LoopedFormer} corresponds to the full-resolution schedule \(\{1,1\}\). For example, \texttt{4+8×\{$1,1$\}+4} is a recursive model with a 4-layer pre-loop block, an 8-layer loop-shared core executed twice at full resolution, and a 4-layer post-loop block, while \texttt{4+8×\{$\frac{1}{8}$,$\frac{1}{4}$,$\frac{1}{2}$,1\}+4} executes the same shared core over a coarse-to-fine schedule.

% \paragraph{SpiralFormer variants (parameter vs.\ compute controls).}
% To disentangle \emph{parameter} and \emph{compute} effects, we report two SpiralFormer variants with complementary controls. \textbf{SpiralFormer-B} (\emph{base}) keeps the same \((N_{\mathrm{pre}},N_{\mathrm{core}},N_{\mathrm{post}})\) layer allocation as \textsc{LoopedFormer}, but replaces the full-resolution schedule \(\{1,1\}\) with a coarse-to-fine schedule \(\{\tfrac{1}{8},\tfrac{1}{4},\tfrac{1}{2},1\}\), which approximates the compute of two full-resolution loops while encouraging global-to-local refinement. \textbf{SpiralFormer-L} (\emph{large}) matches the parameter count of the non-recursive baseline by allocating more layers to the untied prelude/coda, and replaces the middle full-resolution computation with a deeper coarse-to-fine schedule \(\{\tfrac{1}{16},\tfrac{1}{8},\tfrac{1}{4},\tfrac{1}{2}\}\), reducing compute under comparable parameters. 
% Notably, SpiralFormer-L has similar compute to SpiralFormer-B, while its parameter budget matches the non-recursive baseline.

\subsection{Main results.}
Table~\ref{tab:main_results} compares SpiralFormer against (i) full-resolution recursive baselines (\textsc{LoopedFormer}) and (ii) standard Pythia (\textsc{Baseline}).
To disentangle compute and parameter effects, we report two variants: \textsc{SpiralFormer-B} (\emph{base}), which keeps the same recursive layer allocation as \textsc{LoopedFormer} but uses a coarse-to-fine multi-resolution schedule, and \textsc{SpiralFormer-L} (\emph{large}), which matches the parameter count of the non-recursive baseline while replaces the middle full-resolution computation with coarse-to-fine recursions.

\textbf{Multi-resolution recursion improves recursive baselines.}
Under the same layer allocation, \textsc{SpiralFormer-B} consistently improves perplexity and few-shot accuracy over full-resolution \textsc{LoopedFormer}, while also reducing FLOPs.
Across sizes, \textsc{SpiralFormer-B} reduces FLOPs by \textbf{$\sim$7--11\%} relative to \textsc{LoopedFormer} (e.g., 4.59$\rightarrow$4.11 at 410M, 9.67$\rightarrow$8.95 at 1B), while maintaining or improving downstream accuracy, which demonstrates that multi-resolution recursion strengthens looped Transformers without increasing parameters.

\textbf{Surpassing non-recursive baselines with fewer compute.}
\textsc{SpiralFormer-L} matches the parameter count of Pythia but reduces FLOPs by executing most loop computation on compressed sequences, while improving quality.
Across sizes, \textsc{SpiralFormer-L} reduces FLOPs by \textbf{$\sim$3--10\%} at matched parameters (e.g., 1.65$\rightarrow$1.49 at 160M, 4.59$\rightarrow$4.16 at 410M, 14.08$\rightarrow$13.13 at 1.4B), and improves perplexity and few-shot performance.
At 1.4B, it reduces FLOPs (14.08$\rightarrow$13.13) while improving 5-shot accuracy (51.93$\rightarrow$54.37).
Overall, coarse-to-fine recursion improves capability per compute and serves as a drop-in efficiency upgrade to standard Transformer scaling.

\subsection{Scaling Effects}
\label{sec:exp-scaling}

\begin{figure}[t]
  \centering
  \includegraphics[width=\linewidth,page=4]{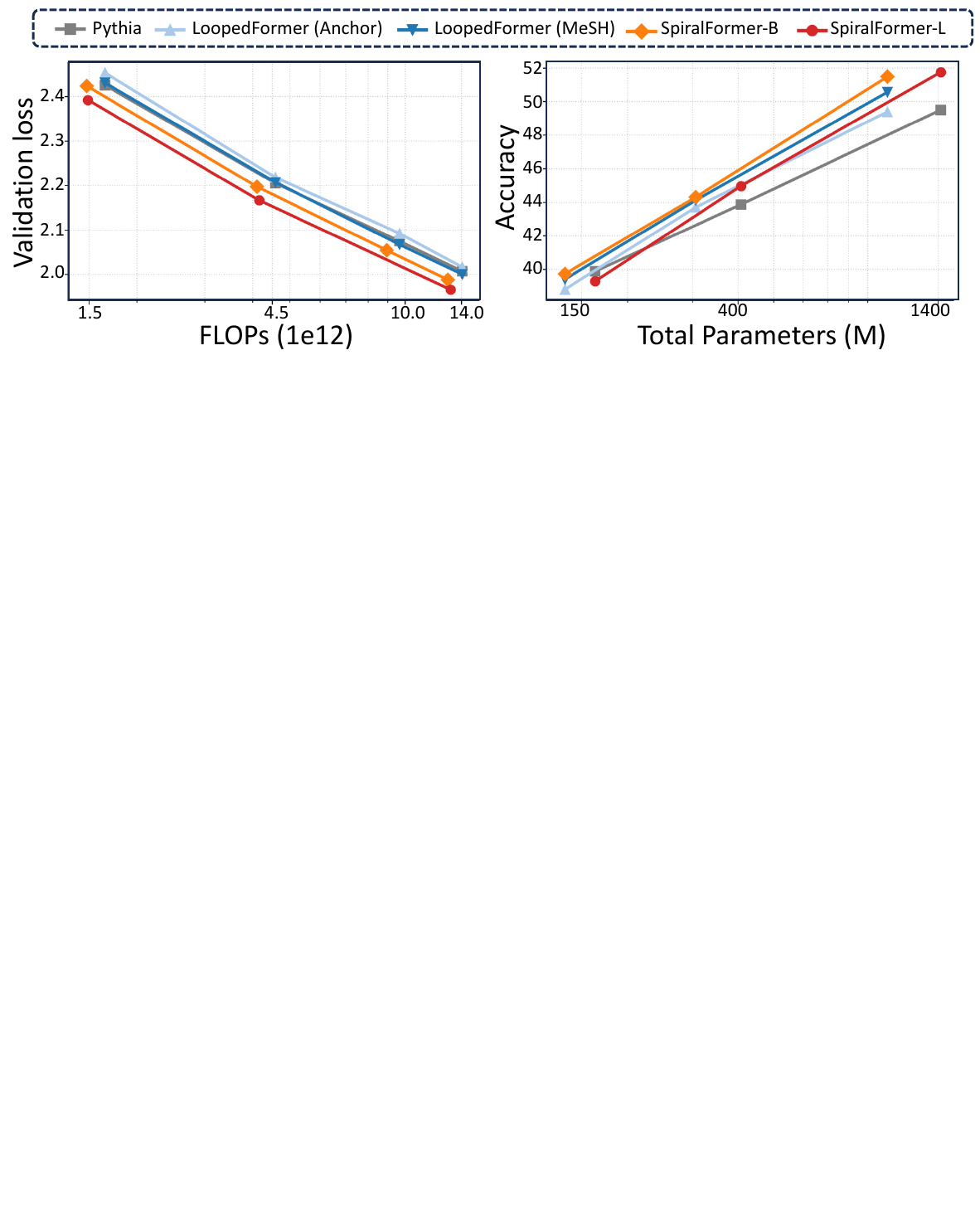}
  \vspace{-225pt}
  \caption{\textbf{Scaling behavior of SpiralFormer.}
  \textbf{Left:} validation loss versus computing FLOPs. 
  % At matched FLOPs, \textsc{SpiralFormer-B} and \textsc{SpiralFormer-L} achieve lower loss than full-resolution recursive baselines (\textsc{LoopedFormer}) and the non-recursive Pythia models, indicating improved capability per unit compute.
  \textbf{Right:} downstream 0-shot accuracy versus total parameters. 
  % At matched parameter count, SpiralFormer variants match or exceed Pythia and \textsc{LoopedFormer}, suggesting that multi-resolution recursion improves parameter efficiency as models scale.
  }
  \label{fig:scaling_effect}
  \vspace{-20pt}
\end{figure}

% \paragraph{Compute scaling (loss vs.\ FLOPs).}
As shown in Figure~\ref{fig:scaling_effect} (left), both \textsc{SpiralFormer-B} and \textsc{SpiralFormer-L} define a consistently better loss--compute frontier than full-resolution \textsc{LoopedFormer} (Anchor/MeSH) and Pythia. Besides, at matched parameter count, SpiralFormer also matches or exceeds both Pythia and \textsc{LoopedFormer} across scales, with the gap generally widening at larger sizes (Figure~\ref{fig:scaling_effect}, right).
Together, the scaling results suggest that multi-resolution recursion yields improvements that persist under scaling, strengthening the overall scaling frontier in both compute- and parameter-constrained regimes.

\subsection{Impact of Recurrence Ratio}
\label{sec:exp-recurrence-ratio}

\begin{figure}[t]
  \centering
  \includegraphics[width=\linewidth,page=5]{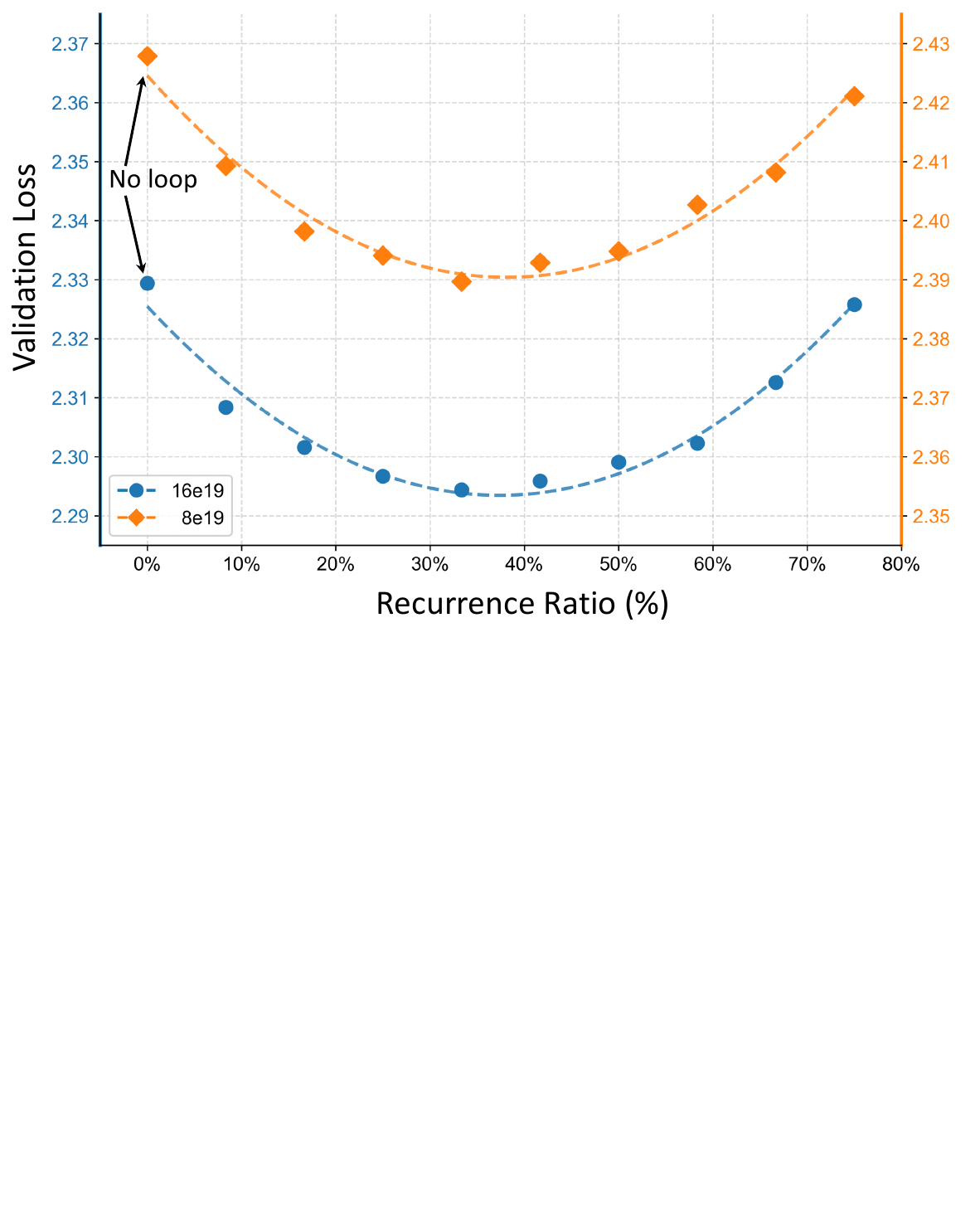} % 替换为你的文件名
  \vspace{-160pt}
  \caption{\textbf{Impact of recurrence ratio on validation loss.}
  We vary the recurrence ratio, defined as the fraction of layers placed in the shared core,
  $N_{\text{loop}}/N_{\text{total}}$, while keeping the total parameter fixed.
  Results at 410m scale are shown under two compute budgets (8e19 and 16e19 FLOPs).
  % Performance is best at a moderate ratio (30--40\%), suggesting an optimal trade-off of recurrence ratio.
  }
  \label{fig:recurrence_ratio_curve}
    \vspace{-20pt}
\end{figure}

To better understand the architectural trade-offs, we study the effect of the
\textbf{Recurrence Ratio} in \textsc{SpiralFormer-L}. 
We define the recurrence ratio as the fraction
\% of layers assigned to the shared core, $N_{\text{loop}}/N_{\text{total}}$, where $N_{\text{total}}$
counts the total number of Transformer layers.
A ratio of 0\% corresponds to a non-recursive MeSH baseline (``No loop''),
while higher ratios indicate more aggressive parameter sharing inside the
multi-resolution recursive core. For all settings we keep
$N_{\text{pre}} = N_{\text{post}}$ in the layer allocation.
We instantiate this study on the 410M-scale model with
$N_{\text{total}} = 24$, and train each configuration at two compute budgets
(8e19 and 16e19 FLOPs). Figure~\ref{fig:recurrence_ratio_curve} plots
validation loss as a function of Recurrence Ratio.

We observe a clear U-shaped curve. Moving from the non-looped baseline
(0\% ratio) to an architecture with even a small amount of looping (10\%)
already yields a substantial improvement in validation loss. As the
Recurrence Ratio increases, the loss continues to decrease and reaches
its minimum around 30–40\%. Beyond this range, further increasing the
ratio leads to degradation, indicating that excessive parameter sharing
starts to hurt capacity.
This pattern is consistent across both compute budgets, suggesting that
the optimal balance between looped and non-looped layers is largely
insensitive to training FLOPs. It supports the choice of
the balanced configuration used for \textsc{SpiralFormer-L} in our main
experiments, and indicates that this balance remains stable as
we scale under different compute regimes.

\begin{table*}[t]
\caption{Ablation study of SpiralFormer components on the 410M model. The first row (highlighted) denotes the \textbf{Main Configuration}. Each subsequent row evaluates the impact of changing \textbf{one} specific design choice from its default setting to a variant. Best results in each column are in \textbf{bold}, and second-best are \underline{underlined}.}
\vspace{-15pt}
\label{tab:ablation_410m}
\vskip 0.15in
\centering
\small
\setlength{\tabcolsep}{5pt} % 恢复正常的列间距
\begin{tabular}{l l cccc cc}
\toprule
\multirow{2}{*}{\textbf{Design Axis}} &
\multirow{2}{*}{\textbf{Ablation (Default $\rightarrow$ Variant)}} &
\multicolumn{4}{c}{\textbf{Perplexity} $\downarrow$} &
\multicolumn{2}{c}{\textbf{Task Acc} $\uparrow$} \\
\cmidrule(lr){3-6} \cmidrule(lr){7-8}
& & Pile & Wiki & LD-O & LD-S & 0-shot & 5-shot \\
\midrule

\rowcolor{gray!15}
\multicolumn{2}{c}{\textbf{Main Configuration (SpiralFormer-B)}}
& \textbf{9.00} & \textbf{21.48} & \underline{19.11} & \textbf{39.78}
& \underline{44.31} & \textbf{46.75} \\
\midrule

(1) Topology & MeSH $\rightarrow$ Anchor & 9.13 & 22.04 & 21.96 & 47.33 & 43.87 & 46.30 \\
(2) Schedule & Coarse-to-Fine $\rightarrow$ Fine-to-Coarse ($\{1,\tfrac{1}{2},\tfrac{1}{4},\tfrac{1}{8}\}$) & 9.24 & 22.38 & 21.56 & 55.36 & 43.61 & 46.01 \\
(3) Causality & Overlap ($s_t=g_t-1$) $\rightarrow$ Parallel ($s_t=g_t$) & 9.10 & 21.86 & 20.91 & 52.37 & 43.53 & 45.70 \\
(4) Offset & Half-chunk ($\omega_t=\lfloor g_t/2 \rfloor$) $\rightarrow$ Zero ($\omega_t=0$) & \underline{9.01} & \underline{21.57} & \textbf{18.69} & \underline{41.29} & \textbf{44.63} & \underline{46.66} \\
\midrule

(5a) Downscale & Self-aggregation $\rightarrow$ Mean pooling & 9.01 & 21.65 & 20.37 & 48.45 & \textbf{44.33} & 46.44 \\
(5b) Upscale & Output-dependent $\rightarrow$ Uniform broadcast & 9.03 & 21.73 & 19.88 & 44.25 & 44.22 & 46.14 \\
(5c) Both Scaling & Learnable $\rightarrow$ Non-learnable (Pooling+Broad.) & 9.03 & 21.64 & 20.13 & 48.86 & 43.97 & 45.40 \\
\bottomrule
\end{tabular}
\vspace{-10pt}
\end{table*}

\subsection{Ablation study.}
We conduct an ablation study on the Pythia-410M model to validate our core design choices (Table~\ref{tab:ablation_410m}).

\textbf{1) Topology.} Replacing MeSH with the Anchor topology consistently degrades performance, indicating that MeSH-style state management provides additional gains for multi-resolution recursion and that the two are highly compatible. The ability of MeSH to route multi-scale updates into a persistent memory buffer further amplifies the representational capacity of the shared core. 

\textbf{2) Schedule.} Switching the resolution schedule from coarse-to-fine to fine-to-coarse also hurts performance, supporting coarse-to-fine as the recommended design. 

\textbf{3) Causality.} Using the no-overlap ``parallel'' regime ($s_t \ge g_t$) further reduces quality, but it introduces deployment-time pipelining opportunities (Appendix~\ref{sec:appendix-parallel-regime}); closing this quality gap while retaining parallelism remains an open direction. 

\textbf{4) Chunk Offset.} Changing the chunk offset to $\omega_t=0$ performs comparably, suggesting it is a viable option; however, offsets affect the periodic triggering pattern and can induce non-uniform per-token compute (Appendix~\ref{sec:appendix-chunk-offset}), so we use the half-chunk offset $\omega_t=\lfloor g_t/2\rfloor$ as a robust default to avoid confounding comparisons.
% , leaving offset design as future work. 

\textbf{5) Down/Up-scaling.} Ablations on the down/up-scaling operators show that self-aggregation outperforms mean pooling and output-dependent allocation outperforms uniform broadcast, and that the learned downscaling and upscaling are compatible and work best when combined.

\section{Analyzing Hierarchical Dependencies in Multi-Resolution Recursion}
\label{sec:analysis}

Multi-resolution recursion is designed to encourage a coarse-to-fine computation pattern: early loop
iterations operate on compressed sequences to capture global structure, while later iterations refine
token-level details at higher resolution.
If the model leverages this structure, we expect the \emph{effective dependency pattern} (as reflected
by attention statistics) to shift systematically across loops as the resolution increases.
In this section we test this hypothesis using two complementary attention-based probes.

\subsection{Measurements for Cross-Loop Dependency Shifts}
\label{sec:analysis-definition}
Let head identity $(\ell,h)$ denote an attention head inside the loop-shared block at layer $\ell$ and head index $h$ within that layer. 
We analyze attention matrices as a function of loop index $t$ and head
identity $(\ell,h)$.
To capture cross-loop changes in dependency patterns, we use two metrics that emphasize different
aspects of attention behavior.

\paragraph{Key-marginal Entropy.}
For each head \((\ell,h)\) at loop \(t\), let \(A^{(t)}_{\ell,h}\in\mathbb{R}^{L_t\times L_t}\) denote the
causal attention matrix at the loop-native length \(L_t=\lfloor r_tL\rfloor\).
We average over queries to obtain a key-marginal distribution
\[
p^{(t)}_{\ell,h}(k) \propto \frac{1}{L_t}\sum_{q=0}^{L_t-1} A^{(t)}_{\ell,h}(q,k),
\quad k=0,\dots,L_t-1,
\]
and define the normalized entropy
\[
H^{(t)}_{\ell,h}
=
-\sum_{k=0}^{L_t-1} p^{(t)}_{\ell,h}(k)\log p^{(t)}_{\ell,h}(k)\,/\,\log L_t \in [0,1].
\]
Lower entropy indicates more concentrated key usage (selective attention), while higher entropy
indicates more diffuse key usage (broad/global attention).

\paragraph{Local Attention Mass (LAM).}
We quantify localness by measuring how much attention probability mass falls into a
resolution-aligned causal local backward window.
Let \(A^{(t)}_{\ell,h}\in\mathbb{R}^{L_t\times L_t}\) be the attention matrix at loop \(t\).
We set the window size
\begin{equation}
M_t \;=\; \big\lceil \gamma\, r_t \big\rceil,\qquad \gamma=32,
\end{equation}
and define the causal local key window for each query \(q\) as
\[
W_t(q)=\{\,k:\ q-M_t \le k < q\,\}\cap \{0,\dots,L_t-1\}.
\]
The LAM score is
\[
\mathrm{LAM}^{(t)}_{\ell,h}
=
\frac{1}{L_t}\sum_{q=0}^{L_t-1}\sum_{k\in W_t(q)} A^{(t)}_{\ell,h}(q,k).
\]
Higher LAM indicates stronger local refinement behavior, while lower LAM indicates weaker locality.

\paragraph{Dynamic heads.}
To focus on heads that exhibit the clearest cross-loop adaptation, we report results on
\emph{dynamic heads} for each metric: the top 40\% heads ranked by cross-loop range
(\(\Delta H\) for entropy and \(\Delta \mathrm{LAM}\) for LAM).
(We include a head-level specialization analysis as supplementary material in
Appendix~\ref{sec:appendix-head-specialization}.)

\begin{figure*}[t]
  \centering
  \includegraphics[width=0.8\textwidth,page=1]{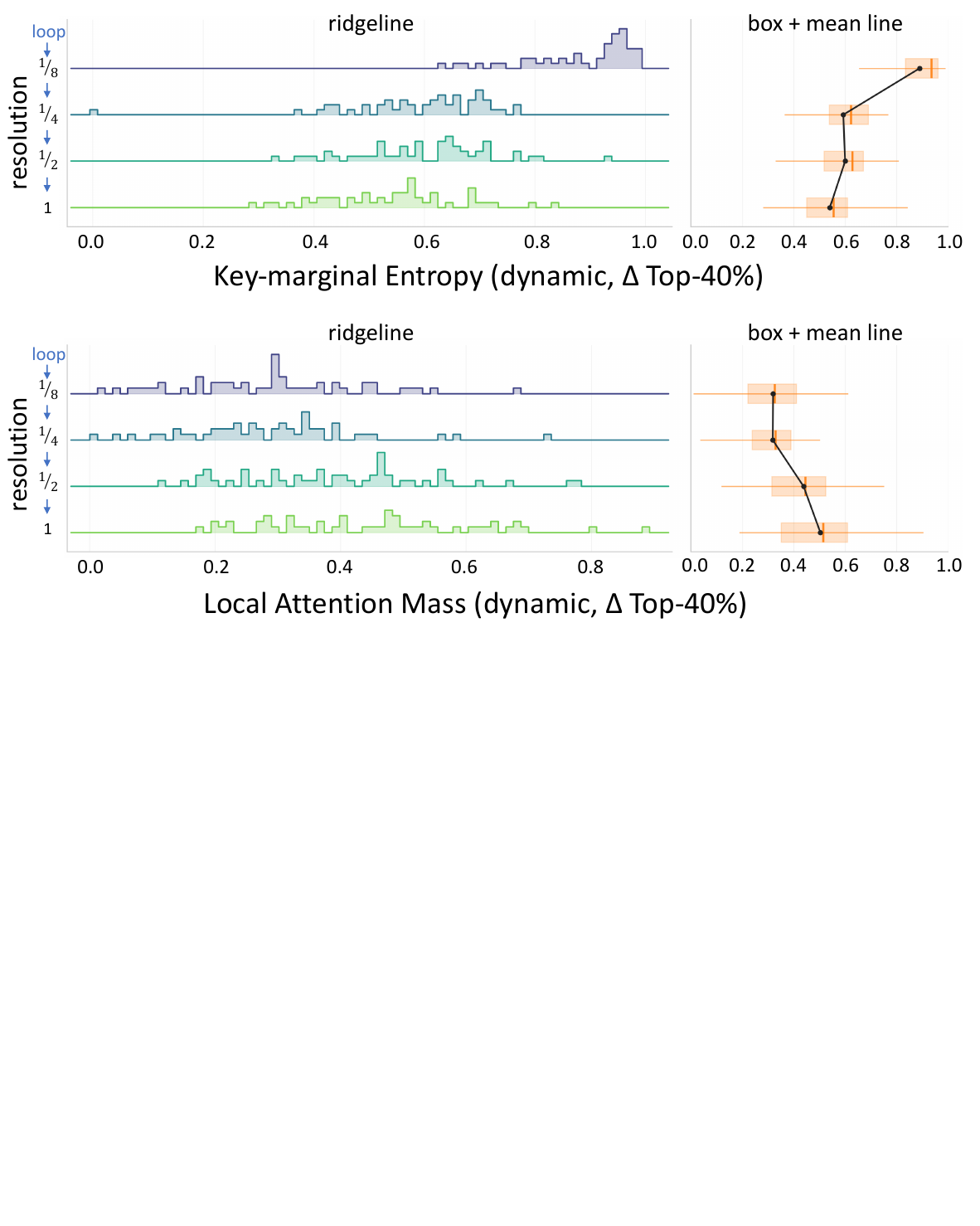}
  \vspace{-245pt}
  \caption{\textbf{Cross-loop distribution shifts of attention statistics on dynamic heads.}
    We visualize the distributions of (top) key-marginal entropy and (bottom) Local Attention Mass
    (LAM) across loop iterations (resolutions) for the \emph{dynamic} heads (top 40\% by cross-loop
    range for each metric) of the 410M \textsc{SpiralFormer-B\meshmark} model.
    Statistics are computed by averaging each head's response over 500 samples from the Pile validation
    set.
    For each loop (from coarse $1/8$ to full resolution $1$), the left subplots show ridgeline
    histograms of head-wise values, and the right subplots show box plots with the loop-wise mean
    indicated by a black dot and connected across loops.
    For completeness,  we further include the distributions over {all heads} in Appendix~\ref{sec:appendix-all-heads}.
}
  \label{fig:analysis_trends}
  \vspace{-14pt}
\end{figure*}

\subsection{Systematic Cross-Loop Shifts Under Coarse-to-Fine Recursion}
\label{sec:analysis-results}

Figure~\ref{fig:analysis_trends} visualizes the distributions of key-marginal entropy (top) and LAM
(bottom) across loop iterations for the 410M \textsc{SpiralFormer-B\meshmark} model under a coarse-to-fine schedule.
Statistics are computed on 500 sequences sampled from the Pile validation set: for each head, we
average the metric over the 500 sequences to obtain one head-wise value per loop, and then plot the
distribution across heads.

\textbf{Observation 1 (attention becomes more selective at higher resolution).}
As resolution increases, the distribution of key-marginal entropy shifts downward
(Figure~\ref{fig:analysis_trends}, top).
This indicates that higher-resolution loops allocate attention mass to a smaller subset of keys,
consistent with a refinement stage that selectively uses a few salient dependencies rather than
maintaining a broad/global interaction pattern.

\textbf{Observation 2 (local refinement strengthens at higher resolution).}
LAM exhibits a consistent cross-loop shift across resolutions
(Figure~\ref{fig:analysis_trends}, bottom).
Interpreting LAM as the fraction of probability mass assigned to local neighborhoods, this provides
evidence that later iterations increasingly emphasize local token-level dependencies, aligning with
the intended role of fine-resolution loops as local refinement steps.

\textbf{Hierarchical dependencies in the loop.}
Together, these two signatures show that the attention behavior changes systematically across loops
as a function of resolution: coarse loops are more diffuse (higher entropy, weaker locality), whereas
fine loops are more selective and more local (lower entropy, higher locality).
This systematic shift supports the claim that multi-resolution recursion induces a hierarchical dependency pattern across iterations---global aggregation at coarse resolution followed by local refinement at fine resolution---even though the core parameters are shared across loops.
We also apply the same probes to a full-resolution \textsc{LoopedFormer\meshmark} baseline and observe much
weaker and less systematic cross-loop shifts (Appendix~\ref{sec:appendix-loopedformer-viz},
Figure~\ref{fig:appendix_loopedformer_viz}), suggesting that the hierarchical pattern is tied to multi-resolution recursion rather than looping alone.

\section{Conclusion}
In this paper, we propose SpiralFormer, a recursive (looped) Transformer architecture that executes loop-shared layers under a multi-resolution recursion schedule. 
% By varying the sequence resolution across loop iterations, SpiralFormer enables a single set of parameters to capture interactions across multiple scales. 
% Our analysis provides both structural and empirical evidence for the effectiveness of the proposed method. 
Empirically, SpiralFormer establishes a superior frontier for both parameter and compute efficiency, consistently outperforming both loop and non-loop baselines across model scales from 160M to 1.4B.
Besides, through attention-based probing, we showed that multi-resolution recursion
% induces systematic, iteration-wise functional specialization, 
enable looped Transformers to learn hierarchical dependencies, opening sequence resolution as a potential axis for scaling recursive architectures.

% \section*{Impact Statements}
% This paper presents work whose goal is to advance the field of machine learning. There are many potential societal consequences of our work, none of which we feel must be specifically highlighted here.


\bibliography{example_paper}
\bibliographystyle{icml2026}

%%%%%%%%%%%%%%%%%%%%%%%%%%%%%%%%%%%%%%%%%%%%%%%%%%%%%%%%%%%%%%%%%%%%%%%%%%%%%%%
%%%%%%%%%%%%%%%%%%%%%%%%%%%%%%%%%%%%%%%%%%%%%%%%%%%%%%%%%%%%%%%%%%%%%%%%%%%%%%%
% APPENDIX
%%%%%%%%%%%%%%%%%%%%%%%%%%%%%%%%%%%%%%%%%%%%%%%%%%%%%%%%%%%%%%%%%%%%%%%%%%%%%%%
%%%%%%%%%%%%%%%%%%%%%%%%%%%%%%%%%%%%%%%%%%%%%%%%%%%%%%%%%%%%%%%%%%%%%%%%%%%%%%%
\newpage
\appendix
\onecolumn

\section{Related Work}
\paragraph{Looped and Recursive Transformers.}
% Looped Transformers decouple computational depth from parameter count by iteratively applying a shared core block \citep{dehghani2018universal}. Recent scaling efforts including Huginn \citep{geiping2025scaling} and Ouro \citep{zhu2025scaling} demonstrate that weight-shared recurrence can match frontier-level performance, while specialized recursion mechanisms including MeSH \citep{yu2025mesh}, Mixture-of-Recursions \citep{bae2025mixture}, Relaxed recursive transformers \citep{bae2024relaxed} and RingFormer \citep{heo2025ringformer} focus on mitigating the representational stagnation observed in looped architectures.
% SpiralFormer provides a complementary perspective by executing the same shared core at varying sequence resolutions, offering a new axis for enhancing the capabilities of looped transformers.
The idea of increasing computational depth by iterating a Transformer block in a loop traces back to the Universal Transformer (UT)~\citep{dehghani2018universal}, which applies a weight-shared layer recurrently and thereby decouples computational depth from parameter depth.
Following UT, a wide array of architectures have explored recursive/looped recurrence in different settings and training regimes~\citep{elbayad2019depth,li2021recurrent,takase2021lessons,shen2022sliced,tan2023sparse,giannou2023looped,yang2023looped,zhang2024autoregressive,fan2024looped,hay2024dynamic,chen2025inner,nguyen2025intra,li2025zero,aleksandrov2025abbie,gong2025makes}.
Recent scaling efforts including Huginn~\citep{geiping2025scaling} and Ouro~\citep{zhu2025scaling} demonstrate that weight-shared recurrence can match frontier-level performance, while specialized recursion mechanisms such as MeSH~\citep{yu2025mesh}, Mixture-of-Recursions~\citep{bae2025mixture}, Relaxed recursive transformers~\citep{bae2024relaxed}, and RingFormer~\citep{heo2025ringformer} further address limitations of vanilla looping (e.g., representational bottlenecks).
SpiralFormer provides a complementary perspective by executing the same shared core at varying sequence resolutions, offering sequence resolution as an additional axis for improving looped Transformers.

\paragraph{Latent Reasoning and Implicit CoT.}
Latent reasoning internalizes the multi-step deliberation of Chain-of-Thought (CoT) within hidden states rather than through explicit text generation \citep{wei2022chain}. Frameworks Continuous Thoughts \citep{hao2024training,shen2025codi}, and Implicit CoT \citep{deng2024explicitcotimplicitcot} enable models to refine internal hypotheses in continuous space before token emission. CCoT \cite{cheng2024compressedchainthoughtefficient} utilizes pretrained models for distillation such that the model only generates important tokens in a continous fashion. SwiReasoning \cite{shi2025swireasoning} saves a large proportion of reasoning tokens by dynamically switching between explicit and implicit decoding mode. A recurring observation in this literature is the \textbf{token efficiency}: complex logic can be compacted into a limited set of high-capacity latent slots to match the performance of much longer explicit reasoning chains. While existing methods primarily focus on distilling latent thoughts into fixed-depth models, SpiralFormer operationalizes this compression morphology as an explicit architectural primitive. By integrating multi-resolution computation into the recursive loop, we formalize the process of global conceptual planning followed by local refinement.

\paragraph{Hierarchical Architectures and Sequence Compression.}
Hierarchical modeling has been extensively explored to alleviate the $O(L^2)$ bottleneck of Transformers by introducing compressed or multi-scale representations. 
ReCAT \citep{hu2023augmenting} augments Transformers with recursively composed multi-grained span representations via contextual inside--outside passes. 
MegaByte \citep{yu2023megabyte}, Hierarchical Autoregressive Transformer \citep{neitemeier2025hierarchicalautoregressivetransformerscombining}, Thought Gestalt \citep{borazjanizadeh2026modelinglanguagesequencethoughts} and Block Transformer \citep{ho2024block} implement global-to-local processing to reduce inference overhead. Patch-level training \citep{shao2024beyond}, MBLM \citep{egli2025multiscalebytelanguagemodels} and H-Net \citep{hwang2025dynamic} pursue training-time efficiency and context-length scaling through token aggregation and learned segmentation. CALM \citep{shao2025continuousautoregressivelanguagemodels} uses an energy-based generative head to predict divergent patch representations. ContextLM \citep{dai2025context} augments next-token prediction with objectives that learn predictive multi-token context embeddings. LCM \citep{barrault2024large} and DLCM \citep{qu2025dynamic} attempt to process text sequence in the concept level.
% Most hierarchical architectures assign different scales to distinct, unshared modules or layer stacks. 
In contrast, SpiralFormer explores a single shared core across a coarse-to-fine resolution schedule within a loop, turning compression into an internal computation primitive rather than a fixed architectural partition. 
% This design encourages the same parameters to operate across scales and, empirically, enables looped Transformers to better learn hierarchical dependencies while retaining the efficiency benefits of multi-scale processing.

\section{MeSH topology: full update equations}
\label{sec:appendix-mesh}

This appendix provides the explicit write--read equations for the MeSH topology used as an
instantiation of the unified topology update in Eq.~\eqref{eq:topo-U}.

\paragraph{Memory buffer.}
MeSH maintains a $B$-slot memory buffer
\[
\bm{M}^{(t)}=\{\bm{m}^{(t)}_b\}_{b=0}^{B-1},\qquad \bm{m}^{(t)}_b\in\mathbb{R}^{L\times d}.
\]
Write/read routers output token-wise distributions over the $B$ slots (softmax over the slot dimension).

\paragraph{Pre-loop initialization ($t=-1$).}
Before the main loop starts, we initialize a designated anchor slot with the input embeddings $\bm{x}$:
\begin{equation}
\bm{m}^{(-1)}_0=\bm{x},\qquad \bm{m}^{(-1)}_{b>0}=\bm{0}.
\label{eq:mesh-init-minus1}
\end{equation}

\paragraph{Transitional write--read (reorganization before the main loop).}
We follow \citet{yu2025mesh} and perform a transitional write--read cycle before entering the main
recursion. Let $\bm{v}=f_{\mathrm{pre}}(\bm{x})$ denote the pre-loop output. Transitional routers take
$\bm{x}$ as input and produce token-wise write/read weights
\begin{equation}
\bm{w}^{(-1)}_{\mathrm{write}}
=\mathrm{Softmax}_b\!\left(\mathrm{Linear}^{(-1)}_{\mathrm{write}}(\bm{x})\right),
\qquad
\bm{w}^{(-1)}_{\mathrm{read}}
=\mathrm{Softmax}_b\!\left(\mathrm{Linear}^{(-1)}_{\mathrm{read}}(\bm{x})\right),
\label{eq:mesh-trans-routers-app}
\end{equation}
where $\bm{w}^{(-1)}_{\mathrm{write}},\bm{w}^{(-1)}_{\mathrm{read}}\in\mathbb{R}^{L\times B}$ and the softmax
is applied over the slot dimension $b$ for each token position.
We write the pre-loop output into the buffer:
\begin{equation}
\bm{m}^{(0)}_{b}
=
\bm{m}^{(-1)}_{b}
+
\bm{v}\odot \bm{w}^{(-1)}_{\mathrm{write},b},
\qquad b=0,\dots,B-1,
\label{eq:mesh-trans-write-app}
\end{equation}
and synthesize the initial loop state by reading from the updated buffer:
\begin{equation}
\bm{h}^{(0)}
=
\sum_{b=0}^{B-1}
\bm{m}^{(0)}_{b}\odot \bm{w}^{(-1)}_{\mathrm{read},b}.
\label{eq:mesh-trans-read-app}
\end{equation}

\paragraph{Main-loop write--read update.}
At iteration $t$, given the current state $\bm{h}^{(t)}$, the causality-corrected update
$\widetilde{\bm{u}}^{(t)}$ (Eq.~\eqref{eq:right-shift}), and the current buffer $\bm{M}^{(t)}$,
MeSH computes step-wise routing weights
\begin{equation}
\bm{w}^{(t)}_{\mathrm{write}}
=\mathrm{Softmax}_b\!\left(\mathrm{Linear}^{(t)}_{\mathrm{write}}(\bm{h}^{(t)})\right),
\qquad
\bm{w}^{(t)}_{\mathrm{read}}
=\mathrm{Softmax}_b\!\left(\mathrm{Linear}^{(t)}_{\mathrm{read}}(\bm{h}^{(t)})\right),
\label{eq:mesh-routers-app}
\end{equation}
updates the buffer by a distributed write
\begin{equation}
\bm{m}^{(t+1)}_{b}
=
\bm{m}^{(t)}_{b}
+
\widetilde{\bm{u}}^{(t)} \odot \bm{w}^{(t)}_{\mathrm{write},b},
\qquad b=0,\dots,B-1,
\label{eq:mesh-write-app}
\end{equation}
and synthesizes the next loop state by reading from the updated buffer:
\begin{equation}
\bm{h}^{(t+1)}
=
\sum_{b=0}^{B-1}
\bm{m}^{(t+1)}_{b}\odot \bm{w}^{(t)}_{\mathrm{read},b}.
\label{eq:mesh-read-app}
\end{equation}
Equations~\eqref{eq:mesh-routers-app}--\eqref{eq:mesh-read-app} define the MeSH instantiation of
the unified topology update in Eq.~\eqref{eq:topo-U} by setting $\mathcal{H}^{(t)}\equiv\bm{M}^{(t)}$.

\section{Right-shift for strict causality and overlap regimes}
\label{sec:appendix-proof-prop1}

\paragraph{Proposition 1 (Causality threshold and overlap).}
Assume at iteration $t$, we partition positions into non-overlapping chunks of size $g_t$. Let $\mathrm{Agg}$ be any within-chunk aggregation operator that maps chunk tokens $\{\bm{h}^{(t)}_i: i\in\mathcal{I}_{t,j}\}$ to a single coarse vector, and let $\mathrm{Exp}$ be any within-chunk expansion operator that maps the coarse vector back to all positions in the same chunk. Denote the pre-shift expanded update by $\bm{u}^{(t)}\in\mathbb{R}^{L\times d}$. We form a shifted update $\widetilde{\bm{u}}^{(t)}$ by right shifting $\bm{u}^{(t)}$ by $s_t$ positions:
\[
\widetilde{\bm{u}}^{(t)}[i]=
\begin{cases}
0,& i<s_t,\\
\bm{u}^{(t)}[i-s_t],& i\ge s_t.
\end{cases}
\]
Then:
(1) If $s_t\le g_t-2$, $\widetilde{\bm{u}}^{(t)}$ is not strictly causal in general.
(2) If $s_t=g_t-1$, $\widetilde{\bm{u}}^{(t)}$ is strictly causal and induces a single-token overlap between the chunk generating the update and the chunk receiving it.
(3) If $s_t\ge g_t$, $\widetilde{\bm{u}}^{(t)}$ is strictly causal and induces no overlap (chunk-wise parallel regime).

\paragraph{Proof.}
Let chunks be indexed by $j=0,\dots,L/g_t-1$. Define the chunk index function $\pi_t(i)=\lfloor i/g_t\rfloor$.
By construction, the pre-shift expansion is chunk-local: for any position $k$,
\begin{equation}
\bm{u}^{(t)}[k] \text{ depends only on } \{\bm{h}^{(t)}_u : u\in \mathcal{I}_{t,\pi_t(k)}\}.
\tag{A.1}
\end{equation}
After shifting, for any $i\ge s_t$, we have $\widetilde{\bm{u}}^{(t)}[i]=\bm{u}^{(t)}[i-s_t]$ and therefore by (A.1),
\begin{equation}
\widetilde{\bm{u}}^{(t)}[i] \text{ depends only on } \{\bm{h}^{(t)}_u : u\in \mathcal{I}_{t,\pi_t(i-s_t)}\}.
\tag{A.2}
\end{equation}

\textbf{(1) Non-causality when $s_t\le g_t-2$.}
Choose any chunk $j\ge 1$ and consider position $i=jg_t+1$ (the second token of chunk $j$). Since $s_t\le g_t-2$, we have $i-s_t \ge jg_t-(g_t-3)$ and there exist valid indices where $\pi_t(i-s_t)=j$. Then by (A.2), $\widetilde{\bm{u}}^{(t)}[i]$ may depend on tokens in chunk $j$, including the last position $jg_t+(g_t-1)$, which is a future token relative to $i$. Hence strict causality can be violated in general.

\textbf{(2) Causality and single-token overlap when $s_t=g_t-1$.}
Fix any $i\ge g_t-1$ and write $i=jg_t+p$ with $p\in\{0,\dots,g_t-1\}$. Then
\[
i-s_t = jg_t+p-(g_t-1) = (j-1)g_t + (p+1).
\]
If $p\le g_t-2$, then $\pi_t(i-s_t)=j-1$, so $\widetilde{\bm{u}}^{(t)}[i]$ depends only on tokens in the previous chunk, all with indices $\le jg_t-1<i$, hence causal.
If $p=g_t-1$, then $i-s_t=jg_t$ and $\pi_t(i-s_t)=j$. This is the unique position in chunk $j$ whose update depends on chunk $j$ itself. Moreover, $\bm{u}^{(t)}[jg_t]$ depends only on tokens within chunk $j$, whose maximum index is $jg_t+(g_t-1)=i$, hence still causal. Therefore strict causality holds and the only overlap is at this single boundary position.

\textbf{(3) Causality and no-overlap when $s_t\ge g_t$.}
For any $i=jg_t+p$, we have $i-s_t\le jg_t+p-g_t\le jg_t-1$, hence $\pi_t(i-s_t)\le j-1$. Therefore $\widetilde{\bm{u}}^{(t)}[i]$ depends only on tokens in chunks strictly before $j$, all with indices $<jg_t\le i$, implying strict causality. Since $\pi_t(i-s_t)\neq j$ for all positions in chunk $j$, there is no overlap between the chunk receiving updates and the chunk generating them. \qed

\section{Inference-Time Motivation for Chunk Offset}
\label{sec:appendix-chunk-offset}

This appendix clarifies an inference-time phenomenon induced by blockwise multi-resolution recursion, and motivates our use of a chunk offset $\omega_t$ in Eq.~\eqref{eq:pi-rho-offset} (default $\omega_t=\lfloor g_t/2\rfloor$).
We refer to Appendix Algorithm~\ref{alg:spiralformer-decode-v3} for pseudocode showing how
$\omega_t$ affects chunk indexing and the resulting chunk-triggered recomputation pattern in decoding.

\paragraph{Phenomenon: periodic triggering leads to non-uniform per-token compute.}
In our multi-resolution recursion, iteration $t$ uses chunk size $g_t$ and a causal right-shift of size $s_t=g_t-1$ (Eq.~\eqref{eq:right-shift}).
Under autoregressive decoding, cached intermediate states can be reused. As a result, low-resolution computations (compress $\rightarrow$ core $\rightarrow$ decompress at chunk size $g_t$) are not necessarily triggered at every token position.
Instead, they are triggered at a periodic subset of positions determined by the chunk structure, which implies that the number of resolution-level computations attributed to different tokens can vary.
Intuitively, tokens that coincide with these triggering positions (one per chunk at each resolution) can incur more work than tokens that do not.
Concretely, in Algorithm~\ref{alg:spiralformer-decode-v3}, low-resolution recomputation at
iteration $t$ is triggered only when the in-chunk index satisfies $\rho=g_t-1$.

\paragraph{Key observation (under $s_t=g_t-1$): one triggering position per chunk.}
Define the offset chunk maps (Eq.~\eqref{eq:pi-rho-offset})
\[
\pi_t(i)=\left\lfloor\frac{i+\omega_t}{g_t}\right\rfloor,\qquad
\rho_t(i)=(i+\omega_t)-g_t\pi_t(i).
\]
Let $\bm{u}^{(t)}$ be the pre-shift upscaled update and $\widetilde{\bm{u}}^{(t)}$ the shifted update.
With $s_t=g_t-1$, we have
\[
\widetilde{\bm{u}}^{(t)}[i] = \bm{u}^{(t)}[i-(g_t-1)].
\]
For positions with $\rho_t(i)\le g_t-2$, the source index $i-(g_t-1)$ lies in the previous chunk, i.e., $\pi_t(i-(g_t-1))=\pi_t(i)-1$.
Therefore, their shifted updates can be formed using already-available summaries from previous chunks.
In contrast, for the unique position with $\rho_t(i)=g_t-1$ in each chunk, the source index satisfies $\pi_t(i-(g_t-1))=\pi_t(i)$, meaning the shifted update can depend on the current chunk summary.
Consequently, at chunk size $g_t$, there is exactly one position per chunk that is responsible for triggering the new low-resolution computation.
This corresponds to the unique ``chunk trigger'' branch in Algorithm~\ref{alg:spiralformer-decode-v3}
(i.e., the $\rho=g_t-1$ condition), which updates the cached latent
$\widehat{\bm{z}}^{(t)}_{\mathrm{last}}$ once per chunk.

\paragraph{Example: non-uniform compute under coarse-to-fine schedules ($\omega_t=0$).}
Consider a coarse-to-fine schedule $[1/8,\,1/4,\,1/2,\,1]$ with chunk sizes $g\in\{8,4,2,1\}$ and $\omega_t=0$ for all iterations.
Then the triggering sets are nested residue classes:
\[
g=2:\ i\equiv 1\ (\mathrm{mod}\ 2),\quad
g=4:\ i\equiv 3\ (\mathrm{mod}\ 4),\quad
g=8:\ i\equiv 7\ (\mathrm{mod}\ 8),
\]
while $g=1$ triggers at every position.
Thus, some tokens (e.g., $i=7$) can trigger multiple resolutions simultaneously, whereas others (e.g., $i=0$) trigger only the full-resolution iteration, yielding non-uniform per-token compute.

\paragraph{How a half-chunk offset changes the triggering pattern.}
With a non-zero offset, the triggering positions shift to a different residue class.
Since $\rho_t(i)=(i+\omega_t)\bmod g_t$, the trigger condition $\rho_t(i)=g_t-1$ is equivalent to
\begin{equation}
i \equiv g_t-1-\omega_t \pmod{g_t}.
\label{eq:trigger-offset-clean}
\end{equation}
Choosing $\omega_t=\lfloor g_t/2\rfloor$ shifts the triggering positions by half a chunk.
Across multiple resolutions (especially when $g_t$ are powers of two), this tends to interleave triggering positions across resolutions, making the per-token compute allocation uniform in practice.
Algorithm~\ref{alg:spiralformer-decode-v3} implements this effect by computing
$(\pi,\rho)$ from $(i+\omega_t)$, so changing $\omega_t$ directly shifts the set of token positions
that satisfy the trigger condition.

\paragraph{Remark.}
Chunk offsets only change how tokens are grouped within each iteration.
They do not affect model parameters, and strict causality is still enforced by the same right-shift operator in Eq.~\eqref{eq:right-shift}.

\section{Inference-Time Parallelism in the Parallel (no-overlap) Regime}
\label{sec:appendix-parallel-regime}

This appendix discusses an inference-time property that arises when using a causal right-shift
size $s_t \ge g_t$ at iteration $t$, i.e., the \emph{no-overlap / parallel regime} in
Proposition~1. While our main experiments set $s_t=g_t-1$ as default, the no-overlap setting reveals an additional form of decoding parallelism that can be exploited in implementations.

\paragraph{Chunk-level time lag induced by $s_t \ge g_t$.}
Consider iteration $t$ with chunk size $g_t$ and shifted update
$\widetilde{\bm{u}}^{(t)}[i] = \bm{u}^{(t)}[i-s_t]$ for $i\ge s_t$.
From Proposition~1(3), when $s_t\ge g_t$, for any position $i$ that lies in chunk $j$,
the source index $i-s_t$ necessarily lies in a chunk strictly earlier than $j$.
Equivalently, the update written into the current chunk cannot depend on any token within the
current chunk.We can summarize this as a \textbf{chunk-level time lag}:
\begin{equation}
\widetilde{\bm{u}}^{(t)}[i] \text{ depends only on }
\{\bm{h}^{(t)}_k:\ \pi_t(k)\le \pi_t(i)-1\},
\qquad \text{when } s_t\ge g_t,
\label{eq:appendix-time-lag}
\end{equation}
where $\pi_t(i)=\lfloor (i+\omega_t)/g_t\rfloor$ is the (offset) chunk index map
defined in Eq.~\eqref{eq:pi-rho-offset}. In particular, the update written into chunk $\pi_t(i)$
cannot depend on any token inside the same chunk, i.e., the multi-resolution recursion is always
at least one chunk behind the full-resolution stream.

\paragraph{Decoding implication: overlapping full-resolution for current token with multi-resolution for future tokens.}
In autoregressive decoding, next-token logits at time step $i$ are computed from the model state at
position $i$. Under the lag property in Eq.~\eqref{eq:appendix-time-lag}, the multi-resolution branch
(with $s_t\ge g_t$) does not require the newly arrived token(s) of the current chunk in order to produce
updates for positions in that chunk; instead it only depends on already-finalized earlier chunks.
This opens a pipeline opportunity:

\begin{itemize}
\item \textbf{Critical path (current token):} compute the full-resolution pathway needed to produce logits
for position $i$ (including pre-/post-loop and any iterations operating at $g=1$).
\item \textbf{Background work (future tokens):} concurrently run the low-resolution recursion
(downscale $\rightarrow$ shared core $\rightarrow$ upscale) on earlier-chunk summaries to precompute
$\widehat{\bm{z}}^{(t)}$ (and/or $\widetilde{\bm{u}}^{(t)}$) that will be consumed by future positions.
\end{itemize}

Concretely, when $s_t \ge g_t$, the low-resolution latent $\widehat{\bm{z}}^{(t)}_j$ for chunk $j$ only depends
on tokens within that chunk and earlier ones. Once chunk $j$ is finalized (i.e., all its fine-grained tokens
have been processed at full resolution), its downscaling and core pass can be launched immediately and
asynchronously, and the resulting $\widehat{\bm{z}}^{(t)}_j$ (or the corresponding shifted updates
$\widetilde{\bm{u}}^{(t)}$) can be cached. Later tokens that fall into chunk $j$ or subsequent chunks then only
need to read these precomputed chunk-level latents on their own critical path. Thus, the bulk of the
multi-resolution computation is moved off the per-token latency path and amortized over time.
In other words, the extra test-time compute introduced by multi-resolution recursion can be scheduled
\emph{off the per-token latency-critical path}, improving hardware utilization while preserving strict
causality.

\paragraph{Connection to cross-loop parallelism in Parallel Loop Transformer.}
The benefit above is closely related in spirit to cross-loop parallelism (CLP) proposed in the
Parallel Loop Transformer (PLT) \citep{wu2025parallel}, which overlaps loop-$\ell$ computation for the
current token with loop-$(\ell+1)$ computation for earlier tokens to collapse sequential loop latency.
Here, the no-overlap regime plays an analogous role: the right-shift $s_t\ge g_t$ removes within-chunk
dependencies, enabling the multi-resolution recursion at iteration $t$ to be advanced on earlier chunks
while the model processes the current token at full resolution.

We view this as a promising systems optimization direction for SpiralFormer deployments, complementary
to the architectural gains from multi-resolution recursion itself.

\section{Implementation Details}
\label{sec:appendix-exp}

\paragraph{Pretraining data.}
All models are pretrained from scratch on a deduplicated subset of the Pile~\citep{gao2020pile} for one epoch (250B tokens), following the Pythia recipe~\citep{biderman2023pythia}. We use the original GPT-NeoX tokenizer with a vocabulary size of 50,257.

\paragraph{Model architecture.}
We use decoder-only GPT-NeoX blocks with RMSNorm and rotary positional embeddings (RoPE). SpiralFormer is instantiated on the Middle-cycle backbone. We denote layer allocations as
\texttt{$N_{\mathrm{pre}}$ + $N_{\mathrm{loop}}\times\{r_0,\dots,r_{T-1}\}$ + $N_{\mathrm{post}}$},
where $N_{\mathrm{loop}}$ layers are shared across $T$ loop iterations and $\{r_t\}$ is the multi-resolution schedule.
Unless stated otherwise, SpiralFormer uses a coarse-to-fine schedule ($r_t=2r_{t-1}$) with chunk sizes $g_t=\lfloor1/r_t\rfloor$, causal right-shift $s_t=g_t-1$ (single-token overlap), and half-chunk offset $\omega_t=\lfloor g_t/2\rfloor$ (Appendix~\ref{sec:appendix-chunk-offset}).
Down/Up-scaling use learnable self-aggregation for downsampling and output-dependent allocation for upsampling (\S\ref{sec:operators}).
We evaluate both Anchor and MeSH topologies; for MeSH, we follow~\citet{yu2025mesh} and treat the causality-corrected update $\widetilde{\mathbf{u}}^{(t)}$ as the produced state written into the MeSH buffer.

\paragraph{Training hyperparameters.}
We use AdamW with $\beta_1=0.9$, $\beta_2=0.95$ and weight decay $0.01$. The learning rate follows a cosine decay schedule with 1\% warmup and decays to 10\% of the peak learning rate. The global batch size is 512 and the sequence length is 4096. All runs use BF16 mixed precision and FlashAttention-2~\citep{dao2023flashattention2}. Distributed training uses DeepSpeed ZeRO Stage 0.

\paragraph{Evaluation protocol.}
We report validation perplexity on the Pile validation slice, WikiText, and Lambada (OpenAI and Standard)~\citep{paperno2016lambada}. For downstream evaluation, we use the LM Evaluation Harness~\citep{eval-harness} and evaluate 0-shot and 5-shot accuracy on 9 tasks: Lambada \citep{paperno2016lambada} in both OpenAI (LD-O) and Standard (LD-S) versions, HellaSwag (HS)~\citep{zellers2019hellaswag}, PIQA~\citep{bisk2020piqa}, WinoGrande (WG)~\citep{sakaguchi2021winogrande}, ARC-Easy (ARC-E) and ARC-Challenge (ARC-C) \citep{clark2018think}, SciQ~\citep{welbl2017crowdsourcing}, and continuation-MMLU (cMMLU)~\citep{hendrycks2020measuring}. We use length-normalized accuracy  for PIQA, HellaSwag, ARC-E, ARC-C, and SciQ and standard accuracy for Lambada, WinoGrande, and cMMLU. All evaluations are conducted in both 0-shot and 5-shot settings, as detailed results are shown in Table~\ref{tab:detail-results}.

\section{Pseudocode}
\label{sec:appendix-pseudocode}
Here we summarize detailed pseudocode for the multi-resolution recursive architecture discussed in the main paper. Algorithm~\ref{alg:mrr} (in the main text) outlines the training-time implementation of SpiralFormer, where a single loop-shared core is executed under a multi-resolution recursion schedule. 
We additionally include Algorithm~\ref{alg:spiralformer-decode-v3} that details the corresponding autoregressive decoding procedure and shows how chunk-triggered multi-resolution updates are combined with caching to realize the same recursion pattern at inference time.
% \newpage
\begin{algorithm}
\caption{Autoregressive decoding with SpiralFormer (chunk-triggered multi-resolution recursion)}
\label{alg:spiralformer-decode-v3}
\begin{algorithmic}[1]
\Require prompt tokens $\{x_0,\dots,x_{n-1}\}$; max new tokens $M$;
chunk sizes $\{g_t\}_{t=0}^{T-1}$; offsets $\{\omega_t\}$; shifts $\{s_t\}$ (default $s_t=g_t-1$);
modules $f_{\mathrm{pre}}, f_{\mathrm{loop}}, f_{\mathrm{post}}$; routers $\{\mathcal{A}^{(t)},\mathcal{B}^{(t)}\}$;
topology update $\mathcal{U}$ (default: MeSH).
\State Init KV caches for $f_{\mathrm{pre}}, f_{\mathrm{loop}}, f_{\mathrm{post}}$
\For{$t=0$ to $T-1$}
    \State Init chunk buffer $\mathbf{B}^{(t)}\in\mathbb{R}^{g_t\times d}$, current chunk id $c^{(t)}\gets -1$
    \State Init last latent $\widehat{\bm{z}}^{(t)}_{\mathrm{last}}\gets \bm{0}$
\EndFor

\State \Call{Prefill}{$\{x_0,\dots,x_{n-1}\}$} \Comment{can be batched over the prompt}
\State $S \gets n$
\For{$m=1$ to $M$} \Comment{Generate (sequential)}
    \State $\bm{h}^{\mathrm{out}} \gets \Call{DecodeStep}{S-1, x_{S-1}}$
    \State $x_S \gets \textsc{Sample}(\textsc{LMHead}(\bm{h}^{\mathrm{out}}))$
    \State $S \gets S+1$
\EndFor

\Function{Prefill}{$\{x_0,\dots,x_{n-1}\}$}
    \For{$i=0$ to $n-1$}
        \State \Call{DecodeStep}{$i, x_i$}
    \EndFor
\EndFunction

\Function{DecodeStep}{$i, x_i$}
    \State $\bm{v}_i \gets f_{\mathrm{pre}}(\mathrm{Embed}(x_i);\ \mathrm{KV}_{\mathrm{pre}})$
    \State $(\bm{h}^{(0)}_i,\mathcal{H}^{(0)}_i) \gets \textsc{TopoInitAt}(i,\mathrm{Embed}(x_i),\bm{v}_i)$ \Comment{e.g., MeSH transitional init}
    \For{$t=0$ to $T-1$}
        \State $(\bm{h}^{(t+1)}_i,\ \mathcal{H}^{(t+1)}_i) \gets \Call{LoopUpdateAt}{t, i, \bm{h}^{(t)}_i,\ \mathcal{H}^{(t)}_i}$
    \EndFor
    \State $\bm{h}^{\mathrm{out}}_i \gets f_{\mathrm{post}}(\bm{h}^{(T)}_i;\ \mathrm{KV}_{\mathrm{post}})$
    \State \Return $\bm{h}^{\mathrm{out}}_i$
\EndFunction

\Function{LoopUpdateAt}{$t, i, \bm{h}^{(t)}_i,\ \mathcal{H}^{(t)}_i$}
    \State $(\pi,\rho)\gets\left(\left\lfloor \frac{i+\omega_t}{g_t}\right\rfloor,\ (i+\omega_t)-g_t\pi\right)$
    \Comment{chunk index and in-chunk offset, Eq.~\eqref{eq:pi-rho-offset}}
    \If{$\pi \neq c^{(t)}$} \Comment{entered a new chunk}
        \State $c^{(t)}\gets \pi$; $\mathbf{B}^{(t)}\gets \mathbf{0}$
    \EndIf
    \State $\mathbf{B}^{(t)}[\rho]\gets \bm{h}^{(t)}_i$
    \Comment{write fine-grained token into chunk buffer}

    \If{$\rho = g_t-1$} \Comment{chunk boundary triggers coarse computation}
        \State $\bm{\alpha}\gets \mathrm{Softmax}_k(\mathcal{A}^{(t)}(\mathbf{B}^{(t)}[k]))$
        \Comment{self-aggregation weights, Eq.~\eqref{eq:self-agg}}
        \State $\bm{z}\gets \sum_{k=0}^{g_t-1}\alpha_k\,\mathbf{B}^{(t)}[k]$
        \Comment{Down-scaling $\mathcal{S}^{(t)}_{\downarrow}$, Eq.~\eqref{eq:downscale-general}}
        \State $\widehat{\bm{z}}^{(t)}_{\mathrm{last}}\gets f_{\mathrm{loop}}(\bm{z};\ \mathrm{KV}_{\mathrm{loop},t})$
        \Comment{shared loop core on compressed sequence, Eq.~\eqref{eq:core-on-z}}
    \EndIf

    \State $\widetilde{\bm{u}} \gets \bm{0}$
    \If{$i \ge s_t$} \Comment{causal right-shift, Eq.~\eqref{eq:right-shift}; default $s_t=g_t-1$ (\S\ref{sec:method-shift})}
        \State $j \gets i-s_t$
        \State $(\pi_s,\rho_s)\gets\left(\left\lfloor \frac{j+\omega_t}{g_t}\right\rfloor,\ (j+\omega_t)-g_t\pi_s\right)$
        \State $\widehat{\bm{z}}_s \gets \widehat{\bm{z}}^{(t)}_{\mathrm{last}}$ \Comment{latest finalized chunk latent}
        \State $\bm{\beta}_s \gets \mathrm{Softmax}(\mathcal{B}^{(t)}(\widehat{\bm{z}}_s))\in\mathbb{R}^{g_t}$
        \Comment{output-dependent allocation, Eq.~\eqref{eq:upscale-router}}
        \State $\widetilde{\bm{u}} \gets \lambda_t\,(\beta_s)_{\rho_s}\cdot \widehat{\bm{z}}_s$  
        \Comment{Up-scaling $\mathcal{S}^{(t)}_{\uparrow}$ with gain $\lambda_t=\sqrt{g_t}$, Eq.~\eqref{eq:upscale-general}}
    \EndIf

    \State $(\bm{h}^{(t+1)}_i,\ \mathcal{H}^{(t+1)}_i)\gets \mathcal{U}(\widetilde{\bm{u}},\ \bm{h}^{(t)}_i,\ \mathcal{H}^{(t)}_i;\ t)$ \Comment{default: MeSH (Eq.~\eqref{eq:mesh-U-compact})}
    \State \Return $(\bm{h}^{(t+1)}_i,\ \mathcal{H}^{(t+1)}_i)$
\EndFunction
\end{algorithmic}
\end{algorithm}

\newpage

\section{Detailed Downstream Results}
\label{sec:appendix-detailed-results}

Table~\ref{tab:detail-results} provides the task-specific 0-shot and 5-shot accuracy for all models across the four scaling points. The average (Avg) column corresponds to the task accuracy reported in the main results (Table~\ref{tab:main_results}).

\begin{table*}[h!]
\caption{Detailed downstream evaluation results on 9 tasks. For each model size, we report 0-shot and 5-shot accuracy. The average accuracy (`Avg') is reported in the final column, corresponding to the values in Table \ref{tab:main_results}. We use notation from the main paper for model configurations, where `\anchormark' denotes Anchor and `\meshmark' denotes MeSH topology. Best result in each column (within a model size and shot setting) is in \textbf{bold}, and second-best is \underline{underlined}.}
\label{tab:detail-results}
\vspace{-10pt}
\vskip 0.1in
\centering
\begin{sc}
\footnotesize
\setlength{\tabcolsep}{4pt} % Adjust column spacing
\resizebox{\textwidth}{!}{%
\begin{tabular}{lll | ccccccccc | c}
\toprule
\textbf{Model} & \textbf{Config} & \textbf{Shot} & \textbf{LD-O} & \textbf{LD-S} & \textbf{HS} & \textbf{PQ} & \textbf{WG} & \textbf{ARC-E} & \textbf{ARC-C} & \textbf{SciQ} & \textbf{MMLU} & \textbf{Avg} \\
\midrule
\multicolumn{13}{l}{\emph{Pythia-160M}} \\
\midrule
\multirow{2}{*}{Baseline (Pythia)} & \multirow{2}{*}{\texttt{12 Layers}} & 0-shot & 32.31 & \underline{23.64} & 31.14 & \textbf{62.46} & 50.59 & \underline{39.56} & 23.21 & \textbf{70.30} & \underline{25.69} & \textbf{39.88} \\
& & 5-shot & \underline{27.11} & \textbf{24.22} & 31.38 & \textbf{62.95} & 50.67 & 42.21 & 22.53 & 78.20 & 25.55 & 40.54 \\
\cmidrule(l){2-13}
\multirow{2}{*}{LoopedFormer\anchormark} & \multirow{2}{*}{\texttt{2+4×\{1,1\}+2}} & 0-shot & 30.04 & 21.11 & 30.93 & 60.39 & \underline{51.14} & 38.13 & \underline{23.81} & 68.30 & 25.40 & 38.81 \\
& & 5-shot & 26.16 & 21.23 & 31.44 & 61.15 & 50.75 & 41.71 & 23.04 & 80.20 & 25.70 & 40.15 \\
\cmidrule(l){2-13}
\multirow{2}{*}{LoopedFormer\meshmark} & \multirow{2}{*}{\texttt{2+4×\{1,1\}+2}} & 0-shot & 31.32 & 21.48 & 31.02 & 60.66 & \textbf{53.43} & 39.06 & 22.27 & \underline{69.70} & \textbf{25.73} & 39.41 \\
& & 5-shot & 26.43 & 21.00 & 31.48 & 60.72 & \underline{51.93} & \underline{42.93} & 23.04 & \underline{81.90} & 26.00 & 40.60 \\
\cmidrule(l){2-13}
\multirow{2}{*}{\textbf{SpiralFormer-B}\meshmark} & \multirow{2}{*}{\texttt{2+4×\{$\tfrac{1}{8}$,$\tfrac{1}{4}$,$\tfrac{1}{2}$,1\}+2}} & 0-shot & \textbf{32.93} & \textbf{23.50} & \underline{31.51} & \underline{61.32} & 50.99 & 39.02 & \textbf{23.98} & 69.10 & 25.27 & \underline{39.73} \\
& & 5-shot & 27.07 & \underline{22.65} & \underline{31.87} & \underline{62.19} & \textbf{52.25} & 42.51 & \underline{23.46} & 81.10 & \underline{26.07} & \underline{41.02} \\
\cmidrule(l){2-13}
\multirow{2}{*}{\textbf{SpiralFormer-L}\meshmark} & \multirow{2}{*}{\texttt{4+4×\{$\tfrac{1}{16}$,$\tfrac{1}{8}$,$\tfrac{1}{4}$,$\tfrac{1}{2}$\}+4}} & 0-shot & \underline{32.78} & 22.76 & \textbf{32.65} & 60.72 & 48.93 & \textbf{40.28} & 23.04 & 67.10 & 25.42 & 39.30 \\
& & 5-shot & \textbf{27.52} & 22.34 & \textbf{32.81} & 61.70 & 51.07 & \textbf{44.36} & \textbf{23.72} & \textbf{82.70} & \textbf{26.09} & \textbf{41.37} \\
\midrule
\multicolumn{13}{l}{\emph{Pythia-410M}} \\
\midrule
\multirow{2}{*}{Baseline (Pythia)} & \multirow{2}{*}{\texttt{24 Layers}} & 0-shot & 41.74 & 29.65 & 37.65 & 64.80 & 51.93 & \underline{43.60} & \textbf{25.68} & 73.10 & 26.68 & 43.87 \\
& & 5-shot & 35.59 & 28.92 & 38.01 & \textbf{67.19} & 50.36 & 50.08 & 25.43 & 85.20 & 27.03 & 45.31 \\
\cmidrule(l){2-13}
\multirow{2}{*}{LoopedFormer\anchormark} & \multirow{2}{*}{\texttt{4+8×\{1,1\}+4}} & 0-shot & 41.45 & 31.24 & 36.82 & 64.09 & 52.80 & 43.14 & 23.72 & \underline{73.60} & 26.39 & 43.70 \\
& & 5-shot & 36.56 & 30.06 & 37.13 & 65.62 & 51.30 & 49.24 & 25.43 & \underline{88.80} & 26.94 & 45.68 \\
\cmidrule(l){2-13}
\multirow{2}{*}{LoopedFormer\meshmark} & \multirow{2}{*}{\texttt{4+8×\{1,1\}+4}} & 0-shot & \underline{41.92} & 32.33 & 37.27 & 64.20 & \underline{53.83} & 42.30 & \underline{25.09} & 73.50 & 26.66 & 44.12 \\
& & 5-shot & 36.25 & 31.71 & 38.00 & 65.40 & 51.22 & 49.37 & 24.49 & 87.00 & 26.93 & 45.56 \\
\cmidrule(l){2-13}
\multirow{2}{*}{\textbf{SpiralFormer-B}\anchormark} & \multirow{2}{*}{\texttt{4+8×\{$\tfrac{1}{8}$,$\tfrac{1}{4}$,$\tfrac{1}{2}$,1\}+4}} & 0-shot & 41.03 & \underline{33.55} & 37.33 & 64.09 & 53.04 & 42.51 & 24.23 & 72.40 & 26.64 & 43.87 \\
& & 5-shot & 36.13 & 32.74 & 38.07 & 65.72 & \underline{52.09} & \underline{50.46} & \textbf{26.11} & 88.10 & \underline{27.30} & 46.30 \\
\cmidrule(l){2-13}
\multirow{2}{*}{\textbf{SpiralFormer-B}\meshmark} & \multirow{2}{*}{\texttt{4+8×\{$\tfrac{1}{8}$,$\tfrac{1}{4}$,$\tfrac{1}{2}$,1\}+4}} & 0-shot & \textbf{43.02} & \textbf{33.69} & \underline{37.98} & \underline{65.83} & 52.49 & 42.72 & 23.64 & 72.70 & \underline{26.72} & \underline{44.31} \\
& & 5-shot & \textbf{38.95} & \textbf{35.94} & \underline{39.00} & 65.56 & \textbf{52.96} & 49.50 & 23.46 & 88.10 & 27.29 & \underline{46.75} \\
\cmidrule(l){2-13}
\multirow{2}{*}{\textbf{SpiralFormer-L}\meshmark} & \multirow{2}{*}{\texttt{8+8×\{$\tfrac{1}{16}$,$\tfrac{1}{8}$,$\tfrac{1}{4}$,$\tfrac{1}{2}$\}+8}} & 0-shot & 41.61 & 32.45 & \textbf{39.63} & \textbf{66.59} & \textbf{53.91} & \textbf{44.91} & 24.57 & \textbf{74.00} & \textbf{27.10} & \textbf{44.97} \\
& & 5-shot & \underline{37.28} & \underline{32.89} & \textbf{40.00} & \underline{66.05} & 51.14 & \textbf{52.65} & \textbf{26.11} & \textbf{89.80} & \textbf{27.63} & \textbf{47.06} \\
\midrule
\multicolumn{13}{l}{\emph{Pythia-1B}} \\
\midrule
\multirow{2}{*}{Baseline (Pythia)} & \multirow{2}{*}{\texttt{16 Layers}} & 0-shot & 46.73 & 34.02 & 43.61 & 66.87 & 52.01 & \textbf{48.53} & \underline{26.28} & 76.60 & 27.86 & 46.95 \\
& & 5-shot & 40.60 & 34.41 & 43.98 & 68.44 & 52.33 & 54.46 & \textbf{28.75} & 89.90 & 28.71 & 49.07 \\
\cmidrule(l){2-13}
\multirow{2}{*}{LoopedFormer\anchormark} & \multirow{2}{*}{\texttt{3+5×\{1,1\}+3}} & 0-shot & 46.17 & 34.68 & 42.62 & \underline{67.68} & \underline{53.51} & 46.80 & 25.26 & 75.90 & \underline{27.99} & 46.73 \\
& & 5-shot & 39.92 & 32.89 & 43.26 & \textbf{69.15} & \underline{53.12} & 55.05 & \underline{27.22} & 90.00 & \underline{28.83} & 48.83 \\
\cmidrule(l){2-13}
\multirow{2}{*}{LoopedFormer\meshmark} & \multirow{2}{*}{\texttt{3+5×\{1,1\}+3}} & 0-shot & \underline{48.40} & \underline{36.95} & \underline{44.36} & 67.03 & 52.01 & 46.93 & \textbf{26.54} & \underline{77.60} & 27.91 & \underline{47.53} \\
& & 5-shot & \underline{42.62} & \underline{34.87} & \underline{44.68} & 67.95 & 52.96 & \underline{55.14} & \underline{27.22} & \underline{91.40} & 28.71 & \underline{49.51} \\
\cmidrule(l){2-13}
\multirow{2}{*}{\textbf{SpiralFormer-B}\meshmark} & \multirow{2}{*}{\texttt{3+5×\{$\tfrac{1}{8}$,$\tfrac{1}{4}$,$\tfrac{1}{2}$,1\}+3}} & 0-shot & \textbf{49.51} & \textbf{37.86} & \textbf{44.85} & \textbf{68.01} & \textbf{55.80} & \underline{47.48} & 23.46 & \textbf{78.10} & \textbf{28.22} & \textbf{48.14} \\
& & 5-shot & \textbf{44.32} & \textbf{37.09} & \textbf{45.71} & \underline{68.83} & \textbf{53.83} & \textbf{55.18} & 26.71 & \textbf{91.60} & \textbf{28.96} & \textbf{50.25} \\
\midrule
\multicolumn{13}{l}{\emph{Pythia-1.4B}} \\
\midrule
\multirow{2}{*}{Baseline (Pythia)} & \multirow{2}{*}{\texttt{24 Layers}} & 0-shot & 51.08 & 39.82 & 47.74 & 68.83 & 55.41 & 50.04 & 26.11 & 77.30 & \underline{29.18} & 49.50 \\
& & 5-shot & 46.17 & 39.69 & 48.01 & 69.64 & 54.22 & 59.22 & 29.27 & 91.20 & 29.95 & 51.93 \\
\cmidrule(l){2-13}
\multirow{2}{*}{Baseline\meshmark} & \multirow{2}{*}{\texttt{24 Layers}} & 0-shot & 53.35 & 42.50 & 49.52 & \underline{69.80} & 53.67 & \textbf{51.77} & 27.90 & 78.50 & 29.05 & 50.67 \\
& & 5-shot & \underline{47.99} & 40.37 & 49.74 & 69.86 & \underline{56.51} & \textbf{61.28} & 30.29 & \underline{93.00} & 30.02 & \underline{53.23} \\
\cmidrule(l){2-13}
\multirow{2}{*}{LoopedFormer\anchormark} & \multirow{2}{*}{\texttt{4+8×\{1,1\}+4}} & 0-shot & 50.75 & 40.93 & 47.65 & 69.75 & 53.75 & 48.40 & 26.62 & 78.00 & 28.62 & 49.39 \\
& & 5-shot & 45.99 & 40.95 & 47.85 & 69.37 & 52.96 & 56.82 & 26.79 & 91.00 & 29.65 & 51.27 \\
\cmidrule(l){2-13}
\multirow{2}{*}{LoopedFormer\meshmark} & \multirow{2}{*}{\texttt{4+8×\{1,1\}+4}} & 0-shot & \underline{53.46} & 41.84 & 48.58 & 69.53 & 54.85 & 49.75 & 27.82 & \underline{80.30} & 28.89 & 50.56 \\
& & 5-shot & \textbf{49.14} & 42.69 & 49.21 & 69.70 & 54.78 & 57.79 & 29.35 & 92.70 & 29.76 & 52.79 \\
\cmidrule(l){2-13}
\multirow{2}{*}{\textbf{SpiralFormer-B}\meshmark} & \multirow{2}{*}{\texttt{4+8×\{$\tfrac{1}{8}$,$\tfrac{1}{4}$,$\tfrac{1}{2}$,1\}+4}} & 0-shot & \textbf{54.03} & \underline{45.29} & \underline{49.57} & \textbf{70.35} & \underline{56.35} & 50.08 & \textbf{29.10} & 79.50 & 29.01 & \underline{51.48} \\
& & 5-shot & 46.61 & \underline{43.82} & \underline{50.51} & \underline{70.08} & 54.30 & 59.68 & \underline{31.49} & 92.30 & \underline{30.15} & 53.22 \\
\cmidrule(l){2-13}
\multirow{2}{*}{\textbf{SpiralFormer-L}\meshmark} & \multirow{2}{*}{\texttt{8+8×\{$\tfrac{1}{16}$,$\tfrac{1}{8}$,$\tfrac{1}{4}$,$\tfrac{1}{2}$\}+8}} & 0-shot & 52.86 & \textbf{46.69} & \textbf{50.88} & 69.64 & \textbf{56.43} & \underline{50.59} & \underline{28.75} & \textbf{80.70} & \textbf{29.24} & \textbf{51.75} \\
& & 5-shot & 46.32 & \textbf{45.29} & \textbf{51.66} & \textbf{70.51} & \textbf{59.51} & \underline{60.61} & \textbf{31.91} & \textbf{93.20} & \textbf{30.35} & \textbf{54.37} \\
\bottomrule
\end{tabular}
}
\end{sc}
\vspace{-20pt}
\end{table*}

\section{More Discussion}
\subsection{Additional Analysis: Where Cross-Loop Changes Concentrate}
\label{sec:appendix-head-specialization}

\paragraph{Definition of dynamic heads (per metric).}
Let $(\ell,h)$ denote an attention head at layer $\ell$ and head index $h$ within that layer.
For a metric $m\in\{H,\mathrm{LAM}\}$, let $m^{(t)}_{\ell,h}$ be the head's metric value at loop $t$
(computed as in \S\ref{sec:analysis-definition}, and averaged over 500 Pile validation sequences as
described above).
We define the cross-loop range
\[
\Delta m_{\ell,h} \;=\; \max_t m^{(t)}_{\ell,h} \;-\; \min_t m^{(t)}_{\ell,h}.
\]
A head $(\ell,h)$ is called \emph{dynamic} under metric $m$ if its $\Delta m_{\ell,h}$ is within the
top 40\% among all heads in the loop-shared block (ranked separately for each metric).

\begin{figure}[h!]
% \vspace{-20pt}
  \centering
  \includegraphics[width=0.7\linewidth,page=2]{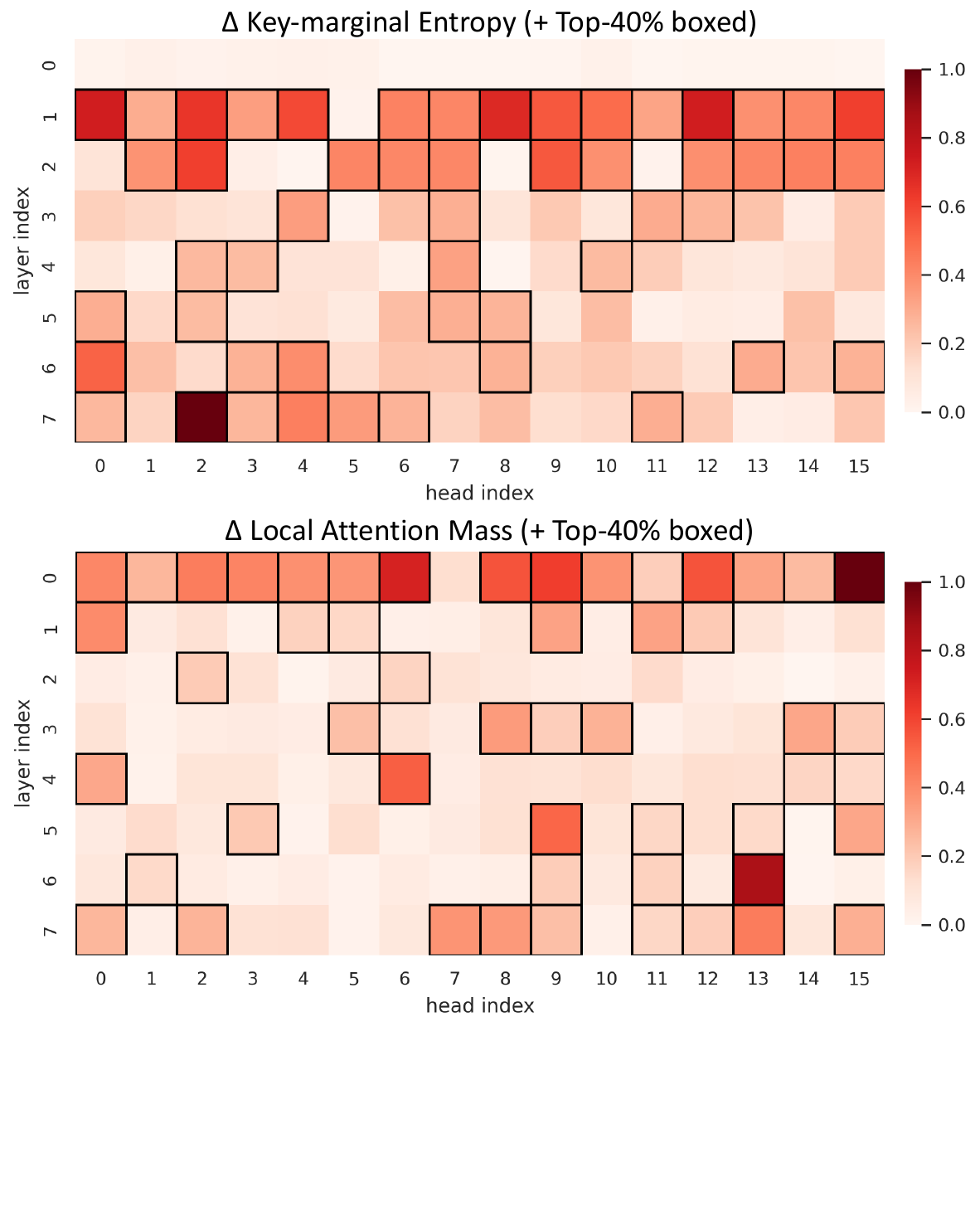}
\vspace{-75pt}
  \caption{\textbf{Head-wise localization of cross-loop variability.}
  We visualize the cross-loop ranges of (top) key-marginal entropy ($\Delta H$) and (bottom) Local
  Attention Mass ($\Delta\mathrm{LAM}$) as layer--head heatmaps, where each cell corresponds to one
  attention head indexed by its layer and head indices $(\ell,h)$.
  To compare patterns across layers/heads, we normalize each heatmap to $[0,1]$ using min--max
  normalization over all cells of that heatmap:
  $x'=(x-\min(x))/(\max(x)-\min(x))$; if $\max(x)-\min(x)$ is numerically negligible (or values are
  non-finite), we set all entries to $0$.
  Black boxes mark \emph{dynamic heads}, defined as the top 40\% heads ranked by cross-loop range for
  the corresponding metric (defined separately per metric).
  Statistics are computed on 500 sequences sampled from the Pile validation set: for each loop $t$
  and head $(\ell,h)$, we first average the metric over the 500 sequences, and then compute cross-loop
  ranges over $t$.}
  \label{fig:appendix_head_heatmap}
  \vspace{-15pt}
\end{figure}

% \paragraph{Observation 3.} Figure~\ref{fig:appendix_head_heatmap} shows that cross-loop changes are not uniformly distributed
% across heads: large variations tend to concentrate in particular layer--head regions.
% Moreover, the highlighted dynamic heads differ between $\Delta H$ and $\Delta\mathrm{LAM}$, indicating
% heterogeneity in how different heads respond to the coarse-to-fine resolution schedule.
% This complements the main-text result that attention statistics shift systematically across loops.
\paragraph{Observation 3 (Cross-loop variability concentrates in a subset of heads).}
Figure~\ref{fig:appendix_head_heatmap} shows that cross-loop changes are not uniformly distributed
across attention heads: large variations in both $\Delta H$ and $\Delta\mathrm{LAM}$ concentrate in
specific layer--head regions.
Moreover, the highlighted dynamic heads differ between $\Delta H$ and $\Delta\mathrm{LAM}$, suggesting
heterogeneity in how heads respond to the coarse-to-fine resolution schedule.
This observation complements the main-text result that attention statistics shift systematically
across loops.

\subsection{Additional Analysis: LoopedFormer Controls (Distribution \& Head-wise)}
\label{sec:appendix-loopedformer-viz}

\textbf{Distribution-level trends.} To contextualize the cross-loop dependency shifts observed in SpiralFormer, we repeat the same
attention-based probes (key-marginal entropy and LAM) on a \emph{full-resolution} LoopedFormer
baseline, where all loop iterations operate at token resolution (i.e., no multi-resolution
compression).
We use the same evaluation setup as in \S\ref{sec:analysis-results}, 500 sequences from the Pile validation set, and report distributions over \emph{dynamic heads} (top 40\% by cross-loop range, defined per metric).

\begin{figure}[h!]
  \centering
  \includegraphics[width=\linewidth,page=7]{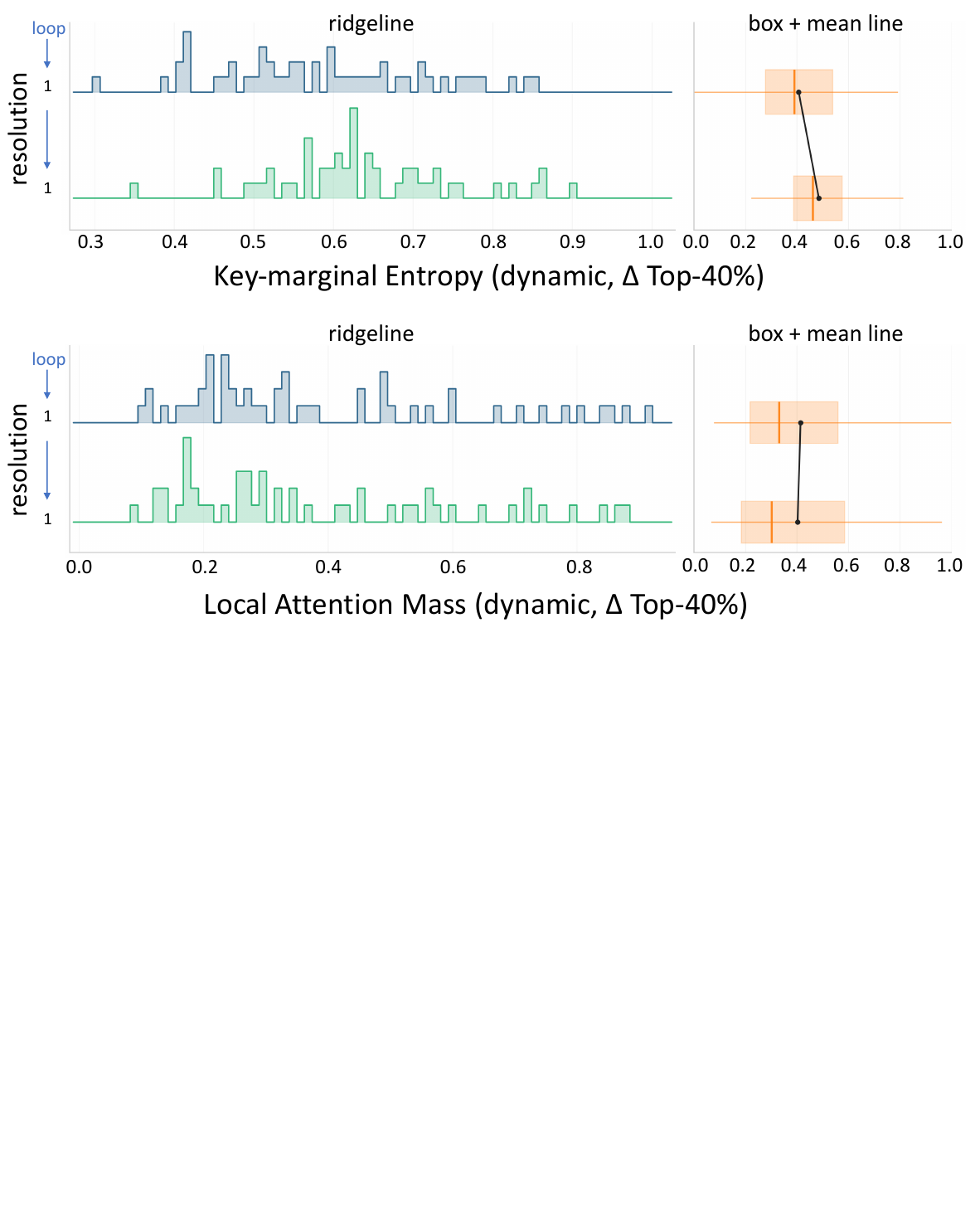}
  \vspace{-300pt}
  \caption{\textbf{LoopedFormer control visualization (full-resolution recursion).}
 Distributions of key-marginal entropy (top) and Local Attention Mass (bottom) across loop iterations
for the 410M full-resolution \textsc{LoopedFormer\meshmark} baseline, under the same probe protocol as Figure~\ref{fig:analysis_trends}.
  We observe qualitatively different cross-loop behaviors compared to SpiralFormer, indicating that resolution changes play an important role in shaping iteration-wise specialization.}
  \label{fig:appendix_loopedformer_viz}
  % \vspace{-10pt}
\end{figure}

\begin{figure}[h!]
  \centering
  % TODO: replace with your actual figure file / page index
  \includegraphics[width=0.7\linewidth,page=9]{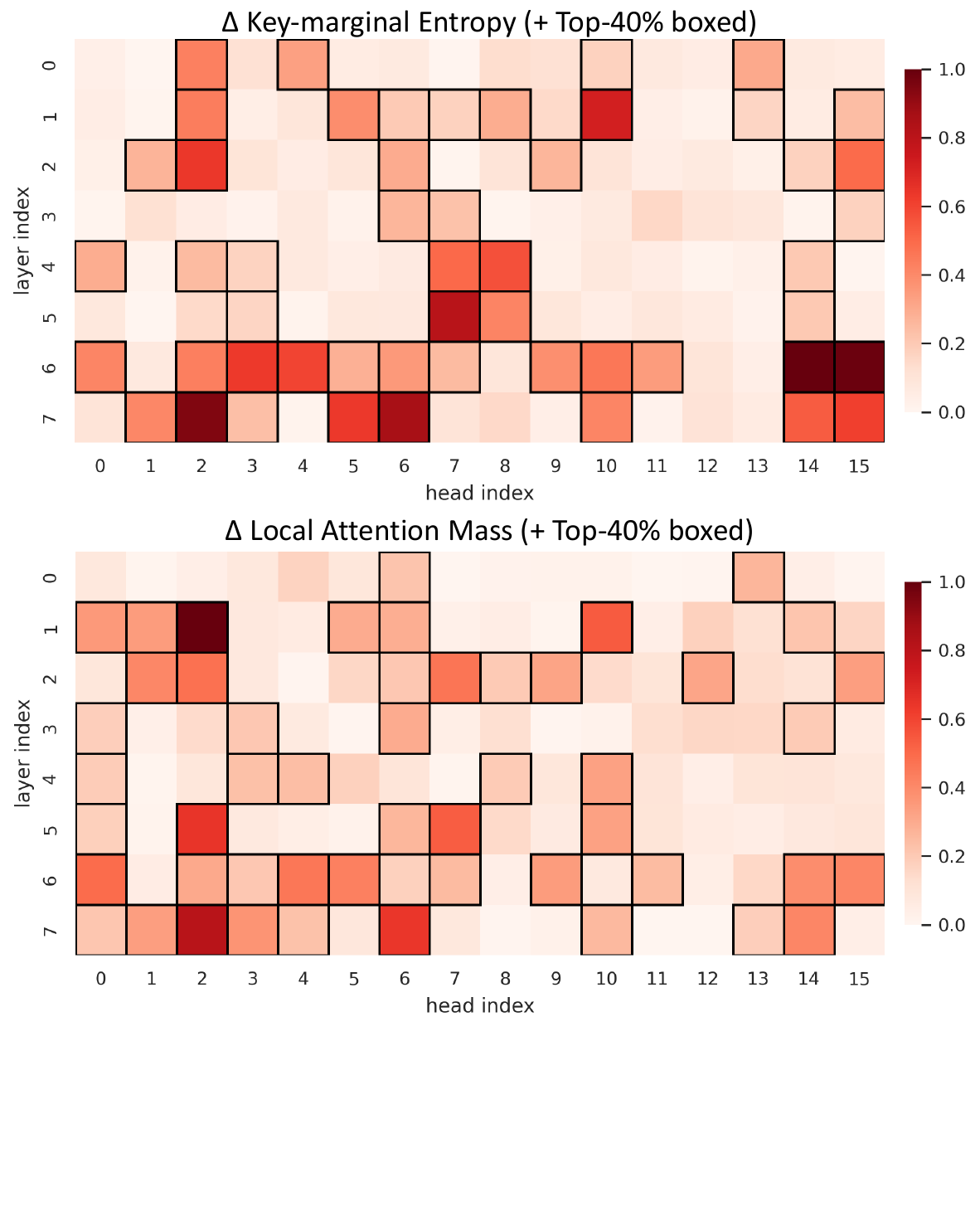}
  \vspace{-65pt}
  \caption{\textbf{Head-wise localization of cross-loop variability in full-resolution
  \textsc{LoopedFormer\meshmark} (control).}
  We visualize the cross-loop ranges of (top) key-marginal entropy ($\Delta H$) and (bottom) Local
  Attention Mass ($\Delta\mathrm{LAM}$) as layer--head heatmaps, with per-heatmap min--max
  normalization to $[0,1]$. Black boxes mark dynamic heads (top 40\% by cross-loop range for each
  metric). Compared to Figure~\ref{fig:appendix_head_heatmap} (SpiralFormer), the control exhibits
  weaker and less structured head-wise specialization.}
  \label{fig:appendix_loopedformer_head_heatmap}
  \vspace{-15pt}
\end{figure}

\newpage
\textbf{Head-wise localization control.}
Beyond distribution-level trends, we additionally examine \emph{where} cross-loop variability
concentrates in the network for a full-resolution \textsc{LoopedFormer\meshmark} control.
We follow the same procedure as in Appendix~\ref{sec:appendix-head-specialization}:
for each head $(\ell,h)$, we compute the cross-loop ranges $\Delta H_{\ell,h}$ and
$\Delta\mathrm{LAM}_{\ell,h}$ (after averaging metrics over 500 Pile validation sequences at each loop),
min--max normalize each heatmap to $[0,1]$, and box dynamic heads (top 40\% by range, per metric).
Figure~\ref{fig:appendix_loopedformer_head_heatmap} shows that \textsc{LoopedFormer} exhibits a
qualitatively different and less structured variability pattern compared to SpiralFormer, supporting
that the iteration-wise specialization observed in the main paper is tied to \emph{multi-resolution}
recursion rather than looping alone.

\newpage

\subsection{Additional Analysis: Attention Statistics over \emph{All} Heads (100\%)}
\label{sec:appendix-all-heads}

\begin{figure*}[h!]
  \centering
  % TODO: replace with your actual figure file / page index
  \includegraphics[width=\textwidth,page=8]{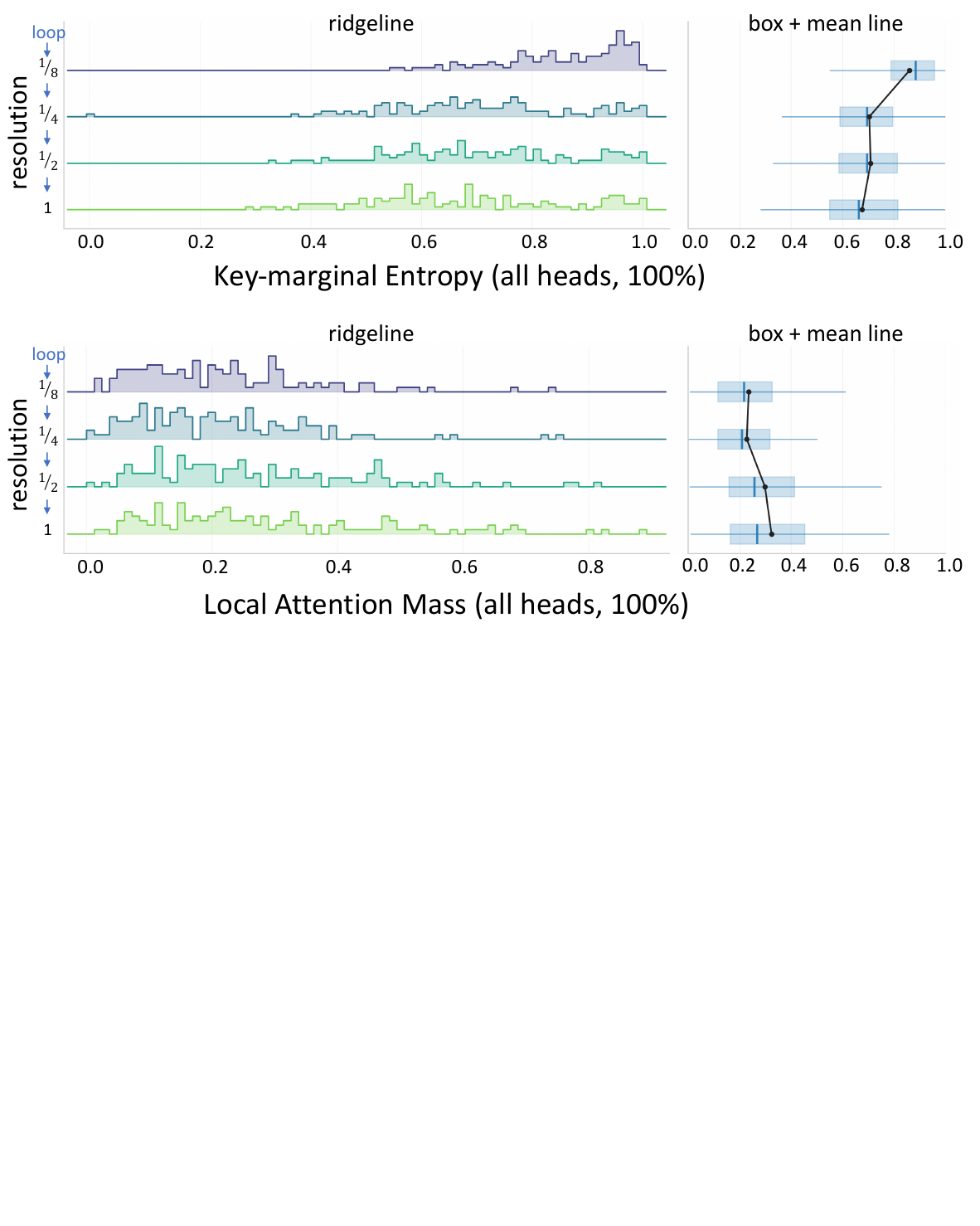}
  \vspace{-300pt}
  \caption{\textbf{Cross-loop distribution shifts of attention statistics on all heads (100\%).}
  Distributions of (top) key-marginal entropy and (bottom) Local Attention Mass (LAM) across loop
  iterations (resolutions) for the \textsc{SpiralFormer-B\meshmark} model at 410M scale.
  Unlike Figure~\ref{fig:analysis_trends}, which focuses on dynamic heads (top 40\% by cross-loop
  range per metric), this figure includes \textbf{all} attention heads in the loop-shared block.
  Statistics are computed by averaging each head's metric over 500 samples from the Pile validation
  set, then plotting the head-wise distributions at each loop. The same coarse-to-fine trend remains
  visible at the population level: entropy decreases and LAM increases as resolution grows.}
  \label{fig:appendix_all_heads}
  \vspace{-10pt}
\end{figure*}

In \S\ref{sec:analysis-results}, to emphasize heads that exhibit the clearest iteration-wise adaptation under
multi-resolution recursion, we reported attention-statistic trends on \emph{dynamic heads} (top 40\%
by cross-loop range; defined separately for key-marginal entropy and Local Attention Mass).
A potential concern is that restricting to dynamic heads may introduce selection bias and overstate
cross-loop specialization. To address this, we provide the same distributional visualizations computed over \textbf{all heads
(100\%)} in the loop-shared block, using the identical evaluation protocol: 500 sequences sampled from
the Pile validation set, and per-head metrics averaged over sequences before forming loop-wise head
distributions. Figure~\ref{fig:appendix_all_heads} shows that the same qualitative cross-loop shifts
remain visible without any head filtering: as resolution increases, (i) key-marginal entropy shifts
downward (attention becomes more selective), and (ii) Local Attention Mass increases (local refinement
strengthens). While the effect sizes are naturally diluted when including heads with weak cross-loop
variation, the persistence of these trends over all heads supports the claim that the observed
hierarchical dependency pattern is a global property induced by multi-resolution recursion rather than
an artifact of selecting a subset of heads.

%%%%%%%%%%%%%%%%%%%%%%%%%%%%%%%%%%%%%%%%%%%%%%%%%%%%%%%%%%%%%%%%%%%%%%%%%%%%%%%
%%%%%%%%%%%%%%%%%%%%%%%%%%%%%%%%%%%%%%%%%%%%%%%%%%%%%%%%%%%%%%%%%%%%%%%%%%%%%%%

\end{document}